\documentclass[10pt]{article} 
\usepackage[accepted]{tmlr}

\usepackage{hyperref}
\usepackage{url}

\usepackage{amsmath,amssymb,amsfonts}
\usepackage{amsthm}
\usepackage{mathtools}
\mathtoolsset{showonlyrefs}
\usepackage{algorithm}
\usepackage[noend]{algpseudocode}
\usepackage{bm}
\usepackage{multirow} 
\usepackage{booktabs} 
\usepackage{xcolor}         
\usepackage{colortbl}
\usepackage{siunitx} 
\newcommand{\best}{\cellcolor{green!15!white}\bf }

\newtheorem{theorem}{Theorem}

\newtheorem{remark}[theorem]{Remark}

\newcommand{\R}{\mathbb{R}}

\newcommand{\E}{\mathbb{E}}

\newcommand{\C}{\mathbb{C}}

\newcommand{\codelocation}{available at \url{https://github.com/johertrich/generative_feature_training}}

\mathtoolsset{showonlyrefs}

\newcommand{\dx}{\mathrm{d}}

\DeclareMathOperator*{\argmin}{arg\,min}

\newcommand{\tT}{\mathrm{T}}

\newcommand{\eps}{\varepsilon}

\title{Generative Feature Training of Thin 2-Layer Networks}

\author{\name Johannes Hertrich \email johannes.hertrich@dauphine.psl.eu \\
      \addr Universit\'e Paris Dauphine - PSL
      \AND
      \name Sebastian Neumayer \email sebastian.neumayer@math.tu-chemnitz.de \\
      \addr Technische Universit\"at Chemnitz
}

\def\month{07}  
\def\year{2025} 
\def\openreview{\url{https://openreview.net/forum?id=6oXNpKuBDK}} 

\begin{document}

\maketitle

\begin{abstract}
    We consider the approximation of functions by 2-layer neural networks with only a few hidden weights based on the squared loss and small datasets.
    Due to the highly non-convex energy landscape, gradient-based training often suffers from local minima.
    As a remedy, we initialize the hidden weights with samples from a learned proposal distribution, which we parameterize as a deep generative model. 
    To train this model, we exploit the fact that with fixed hidden weights, the optimal output  weights solve a linear equation.
    After learning the generative model, we refine a set of sampled weights with a gradient-based feature refinement in the latent space.
    Here, we also include a regularization scheme to counteract potential noise.
    Finally, we demonstrate the effectiveness of our approach by numerical examples.
\end{abstract}

\section{Introduction}
We investigate the approximation of real-valued functions $f\colon [0,1]^d\to\R$ based on samples $(x_k,y_k)_{k=1}^M$, where $x_k\in [0,1]^d$ are independently drawn from some distribution $\nu_{\text{data}}$ and $y_k\approx f(x_k)$ are possibly noisy observations of $f(x_k)$.
To achieve this, we study parametric architectures $f_{w,b}\colon [0,1]^d \to \R$ of the form
\begin{equation}\label{eq:Architecture}
    f_{w,b}(x)=\mathfrak{Re}\biggl(\sum_{l=1}^Nb_l \Phi(\langle w_l,x\rangle)\biggr),
\end{equation}
where $\mathfrak{Re}$ denotes the real part, $\Phi \colon \R \to \C$ is a nonlinear function, and $w_1,...,w_N\in\R^d$ are the features with corresponding weights $b_1,...,b_N\in\C$.
If the function $\Phi$ is real-valued, the model \eqref{eq:Architecture} simplifies to a standard 2-layer neural network architecture without $\mathfrak{Re}$ and with $b_1,...,b_N\in\R$.
The more general model \eqref{eq:Architecture} also covers other frameworks such as random Fourier features \citep{RahRec2007}.
Since the Pareto principle suggests that most real-world systems are driven by a few low-complexity interactions, we are interested in representations \eqref{eq:Architecture} with only a few features $w_l$.
Such an explicit restriction of $N$ also mitigates overfitting, as seen in sparse neural networks, compressed sensing and feature selection.

For fixed $\Phi$ and $N$, we aim to find $(w,b) \in \R^{d,N} \times \C^N$ such that the $f_{w,b}$ from \eqref{eq:Architecture} approximates $f$ well.
From a theoretical perspective, we can obtain such $(\hat w,\hat b) $ by minimizing the mean squared error (MSE), namely
\begin{equation}\label{eq:ContLoss}
     \bigl(\hat w,\hat b\bigr) \in \argmin_{w,b} \|f-f_{w,b}\|^2_{L^2(\nu_{\text{data}})}.
\end{equation}
In practice, we do not have access to $\nu_{\text{data}}$ and $f$, but only to data points $(x_k,y_k)_{k=1}^M$, where $x_k$ are iid samples from $\nu_\text{data}$ and $y_k$ are noisy versions of $f(x_k)$.
Hence, we replace \eqref{eq:ContLoss} by the empirical risk minimization
\begin{equation}\label{eq:EmpRisk}
     \bigl(\hat w,\hat b\bigr) \in \argmin_{w,b} \sum_{k=1}^M |y_k-f_{w,b}(x_k)|^2.
\end{equation}
However, if $M$ is small, minimizing \eqref{eq:EmpRisk} can lead to overfitting towards the training samples $(x_k,y_k)_{k=1}^M$ and poor generalization.
To address this issue, we investigate the following principles.
\begin{itemize}
    \item  We use $f_{w,b}$ of the form \eqref{eq:Architecture} with small $N$.
    This amounts to the implicit assumption that $f$ can be \emph{sparsely} represented using \eqref{eq:Architecture}.
    Unfortunately, under-parameterized networks ($N \ll M$) are difficult to train with conventional gradient-based algorithms \citep{BooDey2022,HolSte2022}, see also Table \ref{tab:results_functions}.
    Hence, we require an alternative training strategy.
    \item Often, we have prior information about the regularity of $f$, i.e., that $f$ is in some Banach space $\mathcal B$ with a norm of the form
    \begin{equation}\label{eq:IntNorm}
        \Vert f \Vert_{\mathcal B}^p = \int_{[0,1]^d} \Vert D f (x) \Vert_q^p \dx x,
    \end{equation}
    where $D$ is some differential operator and $p,q\geq 1$.
    A common example within this framework is the space of bounded variation \citep{Ambrosio2000}, which informally corresponds to the choice $D= \nabla$, $q=2$ and $p=1$.
    In practice, the integral in \eqref{eq:IntNorm} is often approximated using Monte Carlo methods with uniformly distributed samples $(\tilde x_m)_{m=1}^S \subset [0,1]^d$.
    If we use \eqref{eq:IntNorm} as regularizer for $f_{w,b}$, the generalization error can be analyzed in Barron spaces \citep{LiXueYan2022}.
\end{itemize}

\paragraph{Contribution}
We propose a generative modeling approach to solve \eqref{eq:EmpRisk}. 
To this end, we first observe that the minimization with respect to $b$ is a linear least squares problem.
Hence, we can analytically express the optimal $\hat b$ in terms of $w$, which leads to a reduced problem.
Using the implicit function theorem, we compute $\nabla_w \hat b(w)$ and hence the gradient of the reduced objective.
To facilitate its optimization, we replace the deterministic features $w$ with stochastic ones, and optimize over their underlying distribution $p_w$ instead.
We parameterize this distribution as $p_w ={G_\theta}_\#\mathcal N(0,I_d)$ with a deep network $G_\theta \colon \R^d \to \R^d$.
Hence, we coin our approach as \emph{generative feature training}.
Further, we propose to add a Monte Carlo approximation of the norm \eqref{eq:IntNorm} to the reduced objective.
With this regularization, we aim to prevent overfitting.

\section{Related Work}\label{sec:RelatedWork}

\paragraph{Random Features}
Random feature models (RFM) first appeared in the context of kernel approximation \citep{RahRec2007,LiuHuaChe2021}, which enables the fast computation of large kernel sums with certain error bounds, see also \cite{RahRec2008,CorMohTal2010,RudRos2017}.
A similar strategy is pursued by \citet{HZS2006} under the name extreme learning machines.
Sparse RFMs \citep{YenLinLin2014} of the form \eqref{eq:Architecture} with only a few active features can be computed based on $\ell_1$ basis pursuit \citep{HasSchShi2023}.
Since this often leads to suboptimal approximation accuracy, later works by \citet{DoLiWa2022,SaScTr2023,BaiLuZha2024} instead proposed to apply pruning or hard-thresholding algorithms to reduce the size of $w$.
Commonly, the features $w$ are sampled from Gaussian mixtures with diagonal covariances.
Unlike our approach, all these methods begin with a large feature set that has to contain sufficiently many relevant ones.
Towards this strong implicit assumption, \cite{PotSch2021,PW2024} propose to identify the relevant subspaces for the feature proposal based on the ANOVA decomposition. 
Unfortunately, this only works if the features $w$ itself are sparse (few non-zero entries), see Figure~\ref{fig:vis-features}.
Sparse features also enable the fast evaluation of the $f_{w,b}$ from \eqref{eq:Architecture} via the non-equispaced fast Fourier transform \citep{DR1993,PST2001}. 
For kernel approximations, this can be also achieved with slicing methods \citep{H2024,HJQ2024}, which are again closely related to RFMs \citep{RQS2024}.

\paragraph{Adaptive Features}
Besides our work, there are several attempts to design data-adapted proposal distributions $p_w$ for random features \citep{LiTonOgl2019,DNM2025}.
Recently, \citet{BolBurDat2023} proposed to sample the features $w$ in regions where it matters, i.e., based on the available gradient information.
While this allows some adaption, the $w$ still remain fixed after sampling them (a so-called greedy approach).
Towards fully adaptive (Fourier) features $w$, \citet{LiZhaWan2019} propose to alternately solve for the optimal $b$, and to then perform a gradient update for the $w$.
\citet{KamKiePle2020} propose to instead update the $w$ based on a Markov Chain Monte Carlo method.
Unlike our approach, both methods do not incorporate the gradient information of $b$ into the update process of $p_w$.
It is well known that the surrogate alternating updates may perform poorly in certain cases.
Note that learnable features have been also used in the context of positional encoding \citep{LiSiLi2021} and implicit kernel learning \citep{LCMYP2019}.

\paragraph{2-Layer ReLU Networks}
We can interpret 2-layer neural networks as adaptive kernel methods \citep{EMaWu2019}.
Moreover, they have essentially the same generalization error as the RFM.
Several works investigate the learning of the architecture \eqref{eq:Architecture} with $\Phi = \mathrm{ReLU}$ based on a (modified) version of the empirical risk minimization \eqref{eq:EmpRisk}.
Based on convex duality, \citet{PilErg2020} derive a semi-definite program to find a global minimizer of \eqref{eq:EmpRisk}.
A huge drawback is that this method scales exponentially in the dimension $d$.
Later, several accelerations based on convex optimization algorithms have been proposed \citep{MisSahPil2022,BaiGauSoj2023}.
Following a different approach, \citet{Barbu2023} proposed to use an alternating minimization over the parameters $w$ and $b$ that keeps the activation pattern fixed throughout the training.
While this has an improved complexity of $\mathcal O(d^3)$ in $d$, the approach is still restricted to ReLU-like functions $\Phi$.
Moreover, gradient-based optimization of the parameters for a generative (proposal) network such as ours is empirically known to scale very well with $d$.
A discussion of the rich literature on global minimization guarantees in the over-parameterized regime ($N \gg M$) is not within the scope of a sparse architecture \eqref{eq:Architecture}.

\paragraph{Bayesian Networks}
Another approach that samples neural network weights is Bayesian neural networks (BNNs) \citep{Neal2012,JosLagBou2022}.
This allows to capture the uncertainty on the weights in overparameterized architectures.
A fundamental difference 
to our approach and RFMs 
is that we sample the features $(w_l)_{l=1}^N$ independently from the same distribution, while BNNs usually learn a separate one for each $w_l$.
Further, BNNs are usually trained by minimizing an evidence lower bound instead of \eqref{eq:loss_von_theta}, see for example \citep{Graves2011,BluCorKav2015}, which is required to prevent collapsing distributions.

\section{Generative Feature Learning}\label{sec:GenLearn}
Given data points $(x_k,y_k)_{k=1}^M$ with $y_k\approx f(x_k)$ for some underlying $f\colon[0,1]^d\to\R$,
we aim to find the optimal features $w=(w_l)_{l=1}^N\subset \R^{d}$ and weights $b\in\C^N$ such that $f_{w,b}\approx f$, where $f_{w,b}$ is defined in \eqref{eq:Architecture}.
Before we give our approach, we discuss two important instances of the nonlinearity $\Phi \colon \R \to \C$ from the literature.
\hfill
\begin{itemize}
\item
\textbf{Fourier Features}: The choice $\Phi(x)=\mathrm{e}^{2\pi \mathrm{i} x}$ is reasonable if the ground-truth function $f$ can be represented by few Fourier features, e.g., if it is smooth.
As discussed in Section \ref{sec:RelatedWork}, the deployed features $w$ are commonly selected by randomized pruning algorithms.
\item \textbf{2-Layer Neural Network}: For $\Phi\colon \R \to \R$, we can restrict ourselves to
$b \in \R^N$.
Common examples are the ReLU $\Phi(x)=\max(x,0)$ and the sigmoid $\Phi(x)=\frac{\mathrm{e}^x}{1+\mathrm{e}^x}$.
Then, $f_{w,b}$ corresponds to a 2-layer neural network (i.e., with one hidden layer).
Using  the so-called bias trick, we can include a bias into \eqref{eq:Architecture}.
That is, we use padded data-points $(x_k,1) \in \R^{d+1}$ such that the last entry of the feature vectors $w_l \in \R^{d+1}$ can act as bias.
Similarly, an output bias can be included.
\end{itemize}
    
In the following, we outline our procedure for optimizing the parameters $w$ and $b$ for a general $f_{w,b}$ of the form \eqref{eq:Architecture}.
First, we derive an analytic formula for the optimal weights $b$ in the  empirical risk minimization \eqref{eq:EmpRisk} with fixed features $w$.
Then, in the spirit of random Fourier features, we propose to sample the $w$ from a proposal distribution $p_w$, which we learn based on the generative modeling ansatz $p_w ={G_\theta}_\#\mathcal N(0,I_d)$.
As last step, we fine-tune the sampled features $w = G_\theta(z)$ by updating a set of sampled latent features $z$ with the Adam optimizer.
In order to be able to deal with noisy function values $y_k \approx f(x_k)$, we can regularize the approximation $f_{w,b}$ during training.
Our complete approach is summarized in Algorithm \ref{alg:Training}.

\begin{algorithm}[t]
	\begin{algorithmic}[1]
	    \State \textbf{Given:} data $(x_k,y_k)_{k=1}^M$, architecture $f_{w,b}$ as in \eqref{eq:Architecture}, generator $G_\theta$, latent distribution $\eta$
		\While{training $G_\theta$}
			\State sample $N$ latent $z_l \sim \eta$ and set $w = {G_\theta}(z)$
            \State compute optimal $b(w)$ and $\nabla_w b(w)$ based on \eqref{eq:b_von_w}
            \State compute $\nabla_\theta \mathcal L(\theta)$ or $\nabla_\theta \mathcal L_\mathrm{reg}(\theta)$ with automatic differentiation
            \State perform Adam update for $\theta$
            \EndWhile
		\If{GFT-r}
            \While{refining $w$}
                \State set $w = {G_\theta}(z)$
                \State compute optimal $b(w)$ and $\nabla_w b(w)$ based on \eqref{eq:b_von_w}
                \State compute $\nabla_z F(z)$ or $\nabla_z F_\mathrm{reg}(z)$ with automatic differentiation \State perform Adam update for $z$
            \EndWhile
		\EndIf
		\State \textbf{Output:} features $w$ and optimal weights $b(w)$
		\caption{GFT and GFT-r training procedures.}
		\label{alg:Training}
	\end{algorithmic}
\end{algorithm}

\subsection{Computing the Optimal Weights}\label{subsec:optimal_weights}

For fixed $w= (w_l)_{l=1}^N$, any optimal weights $b(w)\in\C^N$ for \eqref{eq:EmpRisk} solve the linear system
\begin{equation}\label{eq:LinSys}
    A_w^\tT A_w b(w)=A_w^\tT y,
\end{equation}
where $y= (y_k)_{k=1}^M$ and $A_w=(\Phi(\langle x_k,w_l\rangle))_{k,l=1}^{N,M}$.
In order to stabilize the numerical solution of \eqref{eq:LinSys}, we deploy Tikhonov regularization with small regularization strength $\eps>0$. This resolves the potential rank deficiency of $A_w^\tT A_w$ and we compute $b(w)$ as the unique solution of
\begin{equation}\label{eq:b_von_w}
(A_w^\tT A_w +\eps I) b(w)=A_w^\tT y.
\end{equation}
For $\epsilon\to0$, the solution of \eqref{eq:b_von_w} converges to the minimal norm solution of \eqref{eq:LinSys}.
A key aspect of our approach is that we can compute $\nabla_w b(w)$ using the implicit function theorem.
This requires solving a linear equation of the form \eqref{eq:b_von_w} with a different right hand side.
For small $N$, the most efficient approach for solving \eqref{eq:b_von_w} is to use a LU decomposition, and to reuse the decomposition for the backward pass.
This procedure is implemented in many automatic differentiation packages such as PyTorch, and no additional coding is required.

By inserting the solution $b(w)$ of \eqref{eq:b_von_w} into the empirical loss \eqref{eq:EmpRisk}, we obtain the reduced loss
\begin{equation}\label{eq:L_von_w}
L(w)=\sum_{k=1}^M \vert f(x_k)-f_{w,b(w)}(x_k) \vert^2.
\end{equation}
Naively, we can try to minimize \eqref{eq:L_von_w} directly via a gradient-based method (such as Adam with its default parameters) starting at some random initialization $w^0 = (w^0_l)_{l=1}^N \subset \R^d$.
 We refer to this as feature optimization (F-Opt).
However, $L(w)$ is non-convex, and our comparisons in Section \ref{sec:Experiments} reveal that feature optimization frequently gets stuck in local minima.
Consequently, a good initialization $w^0$ is crucial if we want to minimize \eqref{eq:L_von_w} with a gradient-based method.
In the spirit of random Fourier features, we propose to initialize the $w$ as independent identically distributed (iid) samples from a proposal distribution $p_w$.
To the best of our knowledge, current random Fourier feature methods all rely on a handcrafted $p_w$.

\subsection{Learning the Proposal Distribution}\label{subsec:GFT}
Since the optimal $p_w$ is in general not expressible without knowledge of $f$, we aim to learn it from the available data $(x_k,y_k)_{k=1}^M$ based on a generative model.
That is, we take a simple latent distribution $\eta$ (such as the normal distribution $\mathcal N(0,I_d)$) and make the parametric ansatz $p_w={G_\theta}_\#\eta$.
Here, $G_\theta\colon\R^d\to\R^d$ is a fully connected neural network with parameters $\theta$ and $\#$ denotes the push-forward of $\eta$ under $G_\theta$.
To optimize the parameters $\theta$ of the distribution $p_w={G_\theta}_\#\eta$, we minimize the expectation of the reduced loss \eqref{eq:L_von_w} with iid features sampled from ${G_\theta}_\#\eta$, namely the loss
\begin{align}
\mathcal L(\theta)= \E_{w\sim ({G_\theta}_\#\eta)^{\otimes N}}\left[L(w)\right]=\E_{z\sim \eta^{\otimes N}}\left[L(G_\theta(z))\right]
= \E_{z\sim \eta^{\otimes N}}\left[\,\sum_{k=1}^M \vert f(x_k)-f_{G_\theta(z),b(G_\theta(z))}(x_k) \vert^2\right]\label{eq:loss_von_theta},
\end{align}
where the notation $\mu^{\otimes N}$ denotes $N$-times the product measure of $\mu$.
We minimize the loss \eqref{eq:loss_von_theta} by a stochastic gradient-based algorithm.
That is, in each step, we sample one realization $z\sim\eta^{\otimes N}$ of the latent features to get an estimate for the expectation in \eqref{eq:loss_von_theta}.
Then, we compute the gradient of the integrand with respect to $\theta$ for this specific $z$, and update $\theta$ with our chosen optimizer.
In the following, we provide some intuition why this outperforms standard training approaches.
In the early training phase, most of the sampled features $w$ do not fit to the data.
Hence, they suffer from vanishing gradients and are updated only slowly.
On the other hand, since the stochastic generator ${G_\theta}_\#\eta$ leads to an evaluation of the objective $L(w)$ at many different locations, we quickly gather gradient information for a large variety of features.
In particular, always taking fresh samples from the iteratively updated proposal distribution $p_w$ helps to get rid of useless features.

\subsection{Feature Refinement: Adam in the Latent Space}\label{subsec:refine}
Once the feature distribution $p_w={G_\theta}_\#\eta$ is learned, we sample a collection $z^0=(z^0_l)_{l=1}^N$ of iid latent features $z^0_l\sim\eta$.
By design, the associated features $w^0= G_\theta(z^0)$ (with $G_\theta$ being applied elementwise to $z^0_1,...,z^0_N$) serve as an estimate for a minimizer of \eqref{eq:L_von_w}. Since these $w^0$ are only an estimate, we refine them similarly as described for the plain feature optimization approach from Section \ref{subsec:optimal_weights}.
More precisely, starting in $z^0$ instead of a random initialization, we minimize the function
\begin{equation}\label{eq:Loss_Latent}
    F(z)=L(G_\theta(z)) = \sum_{k=1}^M \vert f(x_k)-f_{G_\theta(z),b(G_\theta(z))}(x_k) \vert^2,
\end{equation}
where $L$ is the loss function from \eqref{eq:L_von_w}.
By noting that $\nabla F(z)=\nabla G_\theta(z)^\tT\nabla L(G_\theta(z))$, this corresponds to initializing the Adam optimizer for the function $L(w)$ with $w^0 = G_\theta(z^0)$, and to additionally precondition it  by the Jacobian matrix of the generator $G_\theta$.
If the step size is chosen appropriately, we expect that the value of $F(z)$ decreases with the iterations.
Conceptually, our refinement approach is similar to many second-order optimization routines, which also require a good initialization for convergence.

\subsection{Regularization for Noisy Data}
If the number of training points $M$ is small or if the noise on the $y_k \approx f(x_k)$ is strong, minimizing the empirical risk \eqref{eq:EmpRisk} can suffer from overfitting (i.e., the usage of high-frequency features). 
To prevent this, we deploy a regularizer of the form \eqref{eq:IntNorm}.
Choosing $p=q=1$ and $D=\nabla$ in \eqref{eq:IntNorm} leads to the following training problem with (anisotropic) total variation regularization \citep{AcaVog1994,ChanEse2005} 
\begin{equation}\label{eq:EmpRiskReg}
    \hat w  \in \argmin_{w} \sum_{k=1}^M |y_k-f_{w, b(w)}(x_k)|^2 + \lambda R(w),\quad R(w)\coloneqq\int_{[a_{\min},a_{\max}]}\|\nabla f_{w,b(w)}(x)\|_1 \dx x,
\end{equation}
where $\lambda>0$, and $a_{\min}=\min\{x_k:k=1,...,M\}$ and $a_{\max}=\max\{x_k:k=1,...,M\}$ are the entry-wise minimum and maximum of the training data.
For our generative training loss \eqref{eq:loss_von_theta}, adding the regularizer from \eqref{eq:EmpRiskReg} leads to 
\begin{equation}\label{eq:L_reg}
\mathcal L_\mathrm{reg}(\theta)=\E_{w\sim ({G_\theta}_\#\eta)^{\otimes N}}\left[L(w)+\lambda R(w)\right].
\end{equation}
Similarly, we replace the $F$ from \eqref{eq:Loss_Latent} for the feature refinement in the latent space by \begin{equation}\label{eq:F_reg}
    F_\mathrm{reg}(z)=F(z)+\lambda R(G_\theta(z)).
\end{equation}
If we have specific knowledge about the function $f$ that we intend to approximate, then we can apply more restrictive regularizers of the form \eqref{eq:EmpRiskReg}.
As discussed in Section~\ref{sec:RelatedWork}, several RFMs instead regularize the feature selection by enforcing that the features $w_l \in \R^d$ only have a few non-zero entries (sparse features).

\begin{remark}
Given the nature of our numerical examples, we only discussed $f\colon[0,1]^d\to\R$ with data points $y_k\in\R$. 
The extension of our method to multivariate $f\colon[0,1]^d\to\R^n$ is straight forward.
\end{remark}

\section{Experiments}\label{sec:Experiments}

We demonstrate the effectiveness of our method with three numerical examples.
First, we visually inspect the obtained features.
Here, we also check if they recover the correct subspaces. 
Secondly, we benchmark our methods on common test functions from approximation theory, i.e., with a known groundtruth.
Lastly, we target regression on some datasets from the UCI database \citep{uci}.

\subsection{Setup and Comparisons}

For all our experiments, we set up the architecture $f_w{,b}$ in \eqref{eq:Architecture} with $N=100$ features $(w_l)_{l=1}^N$ and one of the nonlinearities $\Phi$ introduced in Section \ref{sec:GenLearn}:
\begin{itemize}
\item We deploy $\Phi(x)=\mathrm{e}^{2\pi\mathrm{i}x}$ without the bias trick.
This corresponds to the approximation of the underlying ground truth function by Fourier features.
\item We deploy $\Phi(x)=\frac{\mathrm{e}^x}{1+\mathrm{e}^x}$, which corresponds to a 2-layer network with sigmoid activation functions.
To improve the expressiveness of the model, we apply the bias trick for both layers.
\end{itemize}
An ablation for different choices of $N$ is given in Appendix~\ref{app:other_N}.
Further, we choose the generator $G_\theta$ for the proposal distribution $p_w = {G_\theta}_{\#} \mathcal N(0,I_d)$ as ReLU network with 3 hidden layers and 512 neurons per hidden layer.
To pick the regularization strength $\lambda$, we divide the original training data into a training (90\%) and a validation (10\%) set.
Then, we train $G_\theta$ for each $\lambda\in\{0\}\cup\{1\times 10^{k}:k=-4,...,0\}$ and choose the $\lambda$ with the best validation error.
To minimize the regularized loss functions $\mathcal L_{\mathrm{reg}}$ (GFT, see also \eqref{eq:L_reg}) and $F_{\mathrm{reg}}$ (GFT-r, see also \eqref{eq:F_reg}), we run $40000$ steps of the Adam optimizer.
The remaining hyperparameters are given in Appendix~\ref{app:impl_details}.
We benchmark all our methods from Section \ref{sec:GenLearn}.
\begin{itemize}
    \item \textbf{F-Opt}: In the feature optimization, we minimize the $L(w)$ from \eqref{eq:L_von_w} with a gradient-based optimizer starting with features $w$ drawn from a standard normal distribution.
    As explained in Section~\ref{subsec:optimal_weights}, we expect that the optimization gets stuck in a local minimum.
    We verify this claim in our experiments.
    \item \textbf{GFT}: For the generative feature training as proposed in Section~\ref{subsec:GFT}, we minimize the loss $\mathcal L(\theta)$ from \eqref{eq:loss_von_theta} and draw iid features $w$ from the generator $G_\theta$ during evaluation.
    \item \textbf{GFT-r}: For the refined generative feature training, we generate features using GFT and refine them with the procedure from Section~\ref{subsec:refine}.
    This requires to minimize the loss $F(z)$ in \eqref{eq:Loss_Latent}.
\end{itemize}
For each method, we specify the choice of $\Phi$ as ``Fourier'' and ``sigmoid'' activation in the corresponding tables.
We compare the obtained results with algorithms from the random Fourier feature literature, and with standard training of neural networks.
More precisely, we consider the following comparisons:
\begin{itemize}
\item \textbf{Sparse Fourier Features}: We compare with the random Fourier feature based methods SHRIMP \citep{DoLiWa2022}, HARFE \citep{SaScTr2023}, SALSA \citep{KanYu2016} and ANOVA-boosted random Fourier features (ANOVA-RFF; \citealp{PW2024}).
We do not rerun the methods and take the results reported by \citet{DoLiWa2022, PW2024}.
\item \textbf{2-Layer Neural Networks}: We train the parameters of the 2-layer neural networks $f_w{,b}$ with the Adam optimizer.
Here, we use exactly the same architecture, loss function and activation function as for GFT.
Additionally, we include results for the ReLU activation function $\Phi(x)=\max(x,0)$.
\item \textbf{Kernel Ridge Regression}: We perform a kernel ridge regression \citep{CS2000} with the Gaussian kernel, where the kernel parameter is chosen by the median rule.
\end{itemize}
Our PyTorch implementation is available online\footnote{\codelocation}.
We run all experiments on a NVIDIA RTX 4090 GPU.
Depending on the specific model, the training takes between 30 seconds and 2 minutes. We include further ablations in Appendix~\ref{app:ablations}.

\subsection{Visualization of Generated Features}

\begin{figure}[hp!]
\centering
\includegraphics[width=.3\textwidth]{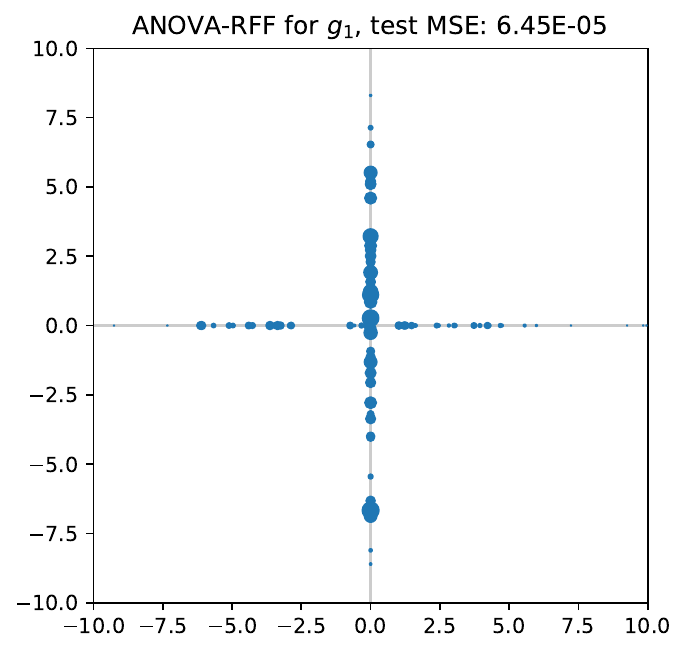}
\includegraphics[width=.3\textwidth]{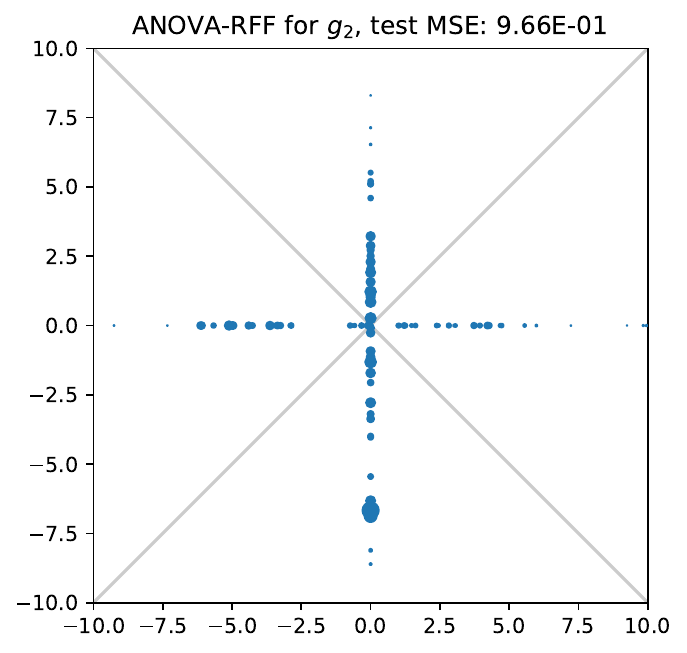}
\includegraphics[width=.3\textwidth]{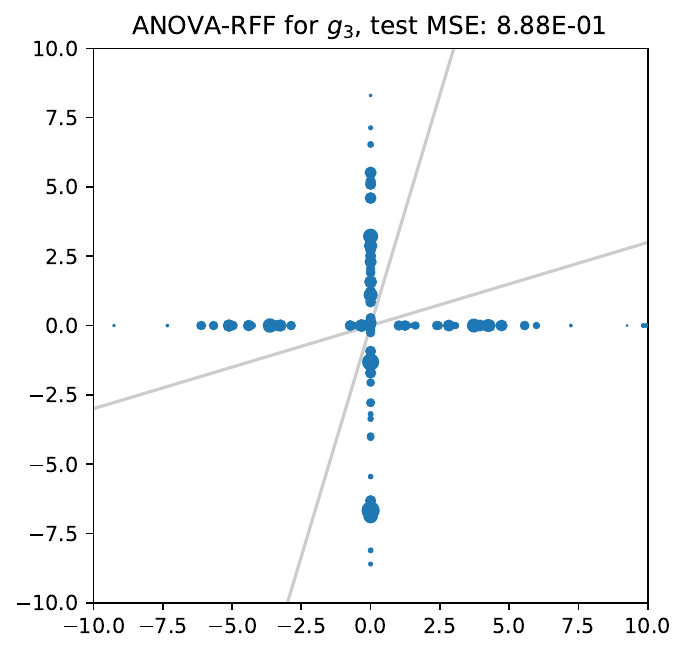}

\includegraphics[width=.3\textwidth]{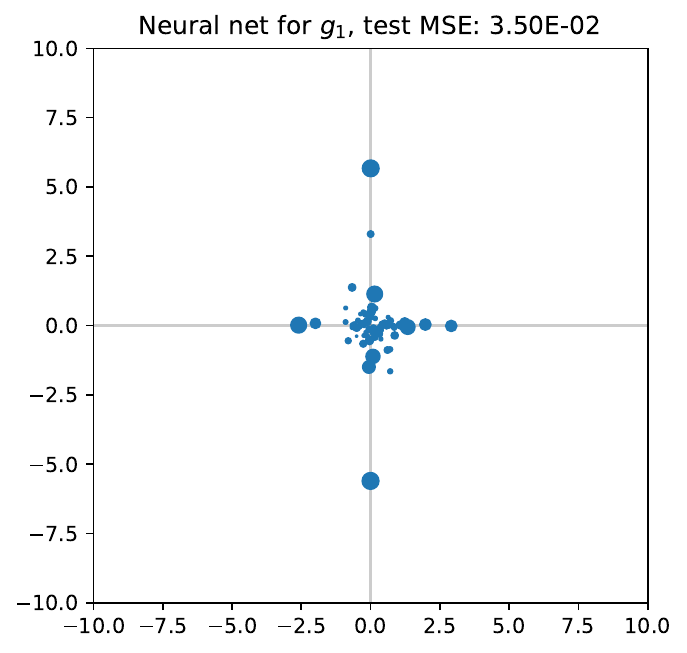}
\includegraphics[width=.3\textwidth]{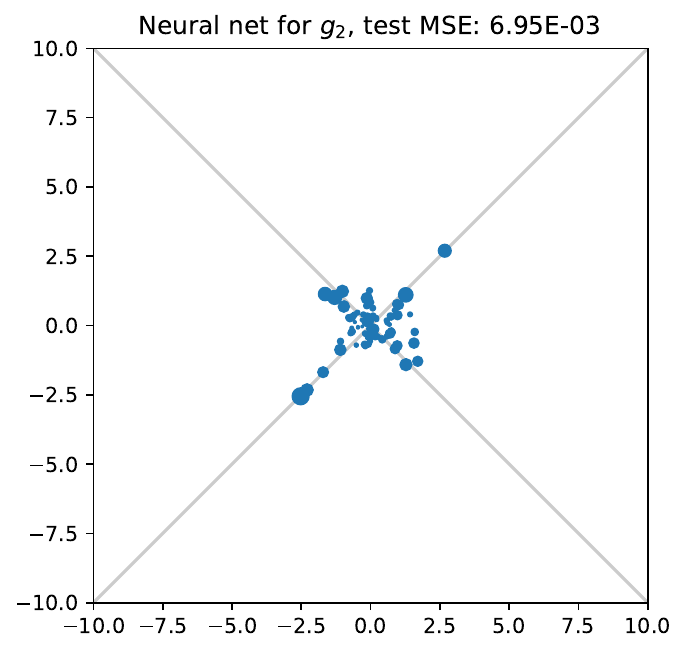}
\includegraphics[width=.3\textwidth]{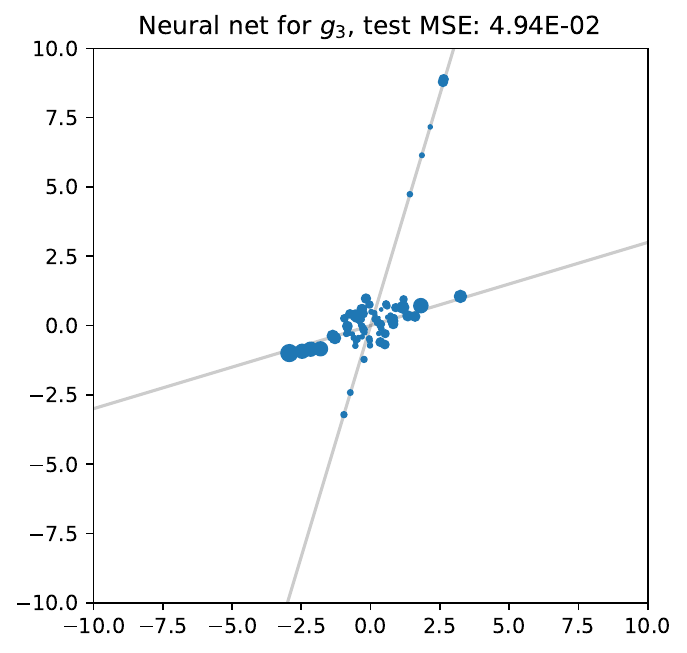}

\includegraphics[width=.3\textwidth]{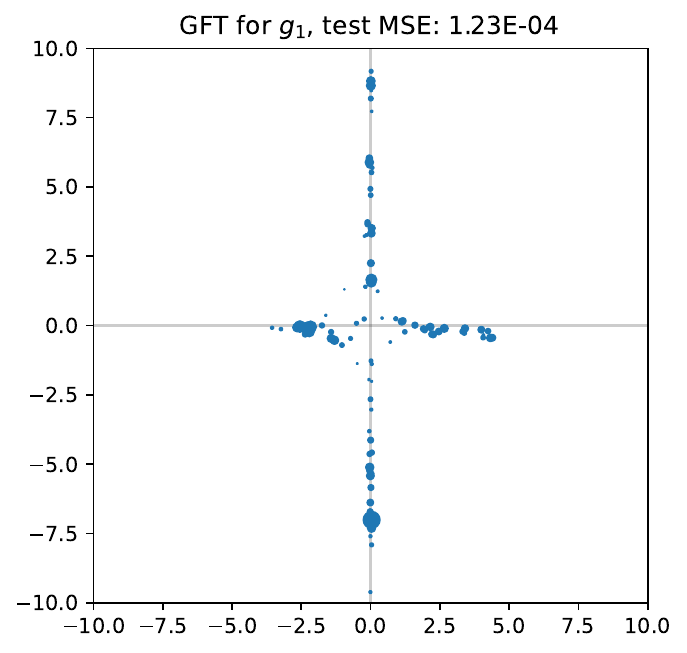}
\includegraphics[width=.3\textwidth]{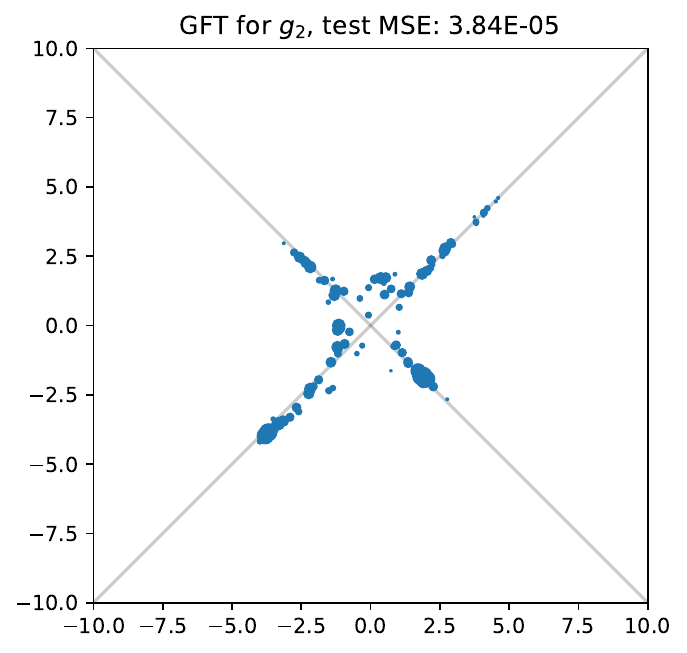}
\includegraphics[width=.3\textwidth]{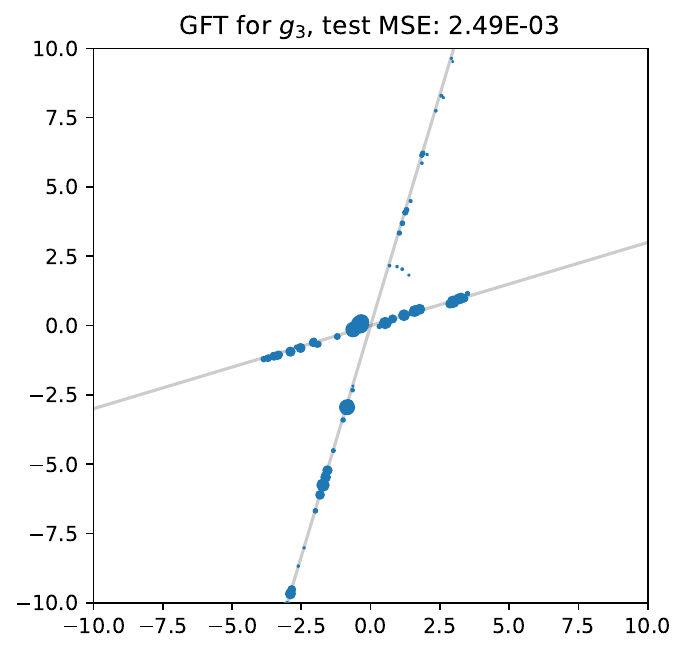}

\includegraphics[width=.3\textwidth]{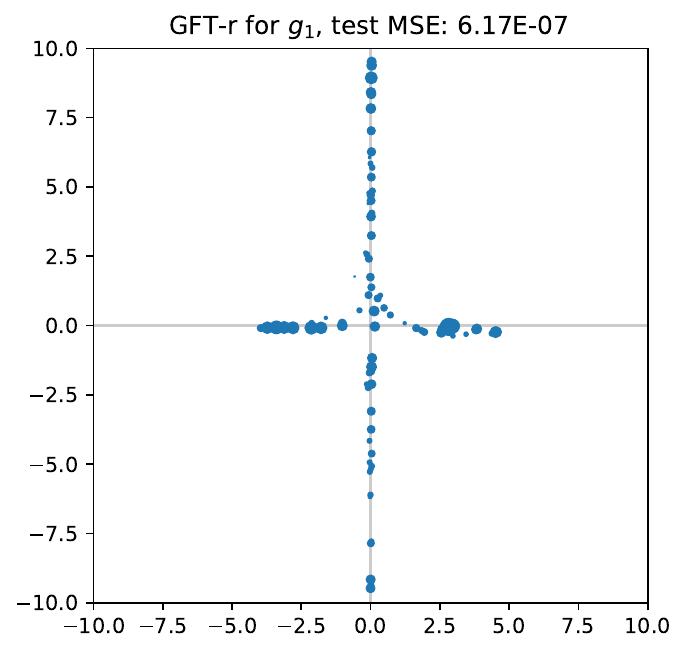}
\includegraphics[width=.3\textwidth]{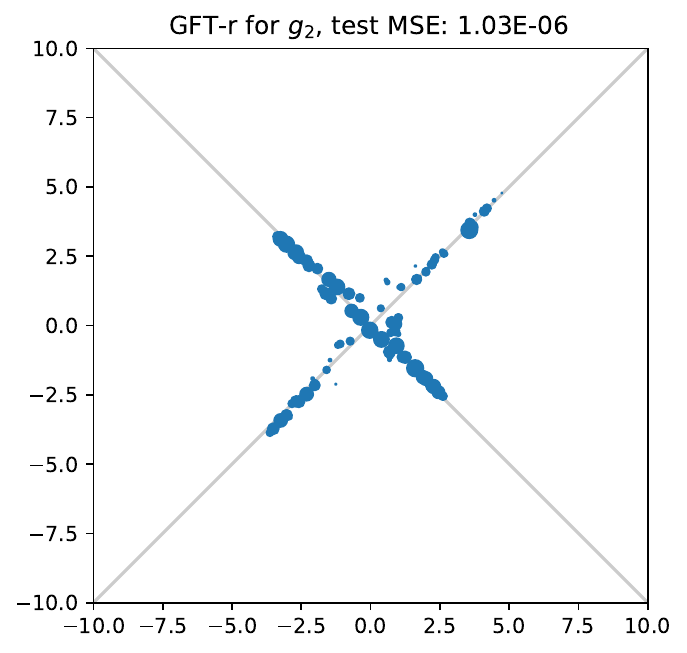}
\includegraphics[width=.3\textwidth]{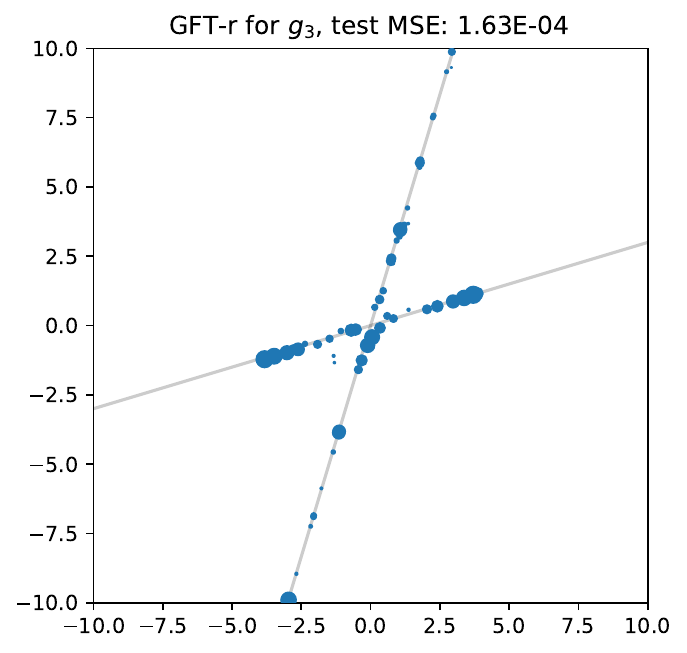}

\caption{Location of the generated features for the $g_i$ from \eqref{eq:vis_funs}, where the marker size reflects the magnitude of the associated weights $b_l$.
The gray lines indicate the support of the Fourier transform of $g_i$.
Since ANOVA-RFF constrains the features to be on the axes, it only works for $g_1$.
For the standard neural network training, the features are not pushed to the axis.
This indicates that the optimization got stuck in a local minimum.}
\label{fig:vis-features}
\end{figure}

First, we inspect the learned features $w$ in a simple setting.
To this end, we consider the function $g\colon\R^2\to\R$ with
$g(x)=\sin(4\pi x_1^2+1)+\cos(4\pi (x_2^4+x_2)$. Since each summand of $g$ depends either on $x_1$ or $x_2$, its Fourier transform is supported on the coordinate axes.
To make the task more challenging, we slightly adapt the problem by concatenating $g$ with two linear transforms $A_i$, which leads to the three test functions
\begin{equation}\label{eq:vis_funs}
g_i(x)=g(A_ix),\quad\text{with}\quad
A_1=\left(\begin{array}{cc}1&0\\0&1\end{array}\right),\quad A_2=\left(\begin{array}{cc}\cos(\frac\pi4)&-\sin(\frac\pi4)\\\sin(\frac\pi4)&\cos(\frac\pi4)\end{array}\right),\quad A_3=\left(\begin{array}{cc}1&0.3\\0.3&1\end{array}\right).
\end{equation}
In all cases, the Fourier transform is supported on a union of two subspaces.
Now, we learn the features $w$ with our GFT and GFT-r method based on $2000$ samples that are drawn uniformly from $[0,1]^2$, and plot them in Figure~\ref{fig:vis-features}. 
The gray lines indicate the support of the Fourier transforms of $g_i$, and the size of the markers indicates the magnitude of the associated $b_l$.
For all functions $g_i$, the features $w$ sampled with GFT are mostly located in the support of the Fourier transform.
The very few features that are located outside of it can be explained by numerical errors and are mostly removed by the refinement procedure GFT-r.
In contrast, for ANOVA-RFF, the $w$ are restricted to be located on the axes. 
Consequently, it cannot work for $g_2$ and $g_3$, and the error obtained is large.
For gradient-based neural network training, the $w$ are not pushed to the axis, indicating that the optimization got stuck at a local minimum.
For functions where the subspaces are orthogonal, such as $g_2$, this issue was recently addressed in \cite{BMWS2024} by learning the associated transform in the feature space.

\subsection{Function Approximation}\label{subsec:function_approx}

We use the same experimental setup as in \citep[Table 7.1]{PW2024}, that is, the test functions
\begin{itemize}
    \item Polynomial: $f_1(x)=x_4^2+x_2x_3+x_1x_2+x_4$;
    \item Isigami: $f_2(x)=\sin(x_1)+7\sin^2(x_2)+0.1 x_3^4\sin(x_1)$;
    \item Friedmann-1: $f_3(x)=10\sin(\pi x_1 x_2)+20(x_3-\frac12)^2+10 x_4+5 x_5$.
\end{itemize}
The input dimension $d$ is set to $5$ or $10$ for each $f_k$.
In particular, the $f_k$ might not depend on all entries of the input $x$.
For their approximation, we are given samples $x_k\sim\mathcal U_{[0,1]^d}$,  $k=1,...,M$, and the corresponding noise-less function values $f_k(x_k)$.
The number of samples $M$ and the dimension $d$ are specified for each setting.
As test set we draw $M$ additional samples from $\mathcal U_{[0,1]^d}$.
We deploy our methods as well as standard neural network training to the architecture $f_{w,b}$.
The MSEs on the test set are given in Table~\ref{tab:results_functions}.
There, we also include ANOVA-random Fourier features, SHRIMP and HARFE for comparison.
We always report the MSE for the best choice of $\rho$ from \citealp[Table 7.1]{PW2024}.
The GFT-r with Fourier activation functions outperforms the other approaches significantly.
In particular, both the GFT and GFT-r consistently improve over the gradient-based training of the architecture $f_{w,b}$.
This is in line with the analysis of gradient-based training in recent works \citep{BooDey2022,HolSte2022}.
As expected, Fourier activation functions are best suited for this task.

\begin{table}[htb]
    \centering
    \caption{Comparison with sparse feature methods for function approximation: We report the MSE over the test set averaged over 5 runs.
    The values for ANOVA-RFF, SHRIMP and HARFE are taken from \citet{PW2024}. 
    The deployed $\lambda$ is indicated below each result. The best performance is highlighted.}
    \vspace{.2cm}
    
   \scalebox{.65}{
\begin{tabular}{cccllllllll}
\toprule
\multicolumn{2}{c}{Method} & \phantom{.}   &  \multicolumn{2}{c}{Function $f_1$}  &  \phantom{.} & \multicolumn{2}{c}{Function $f_2$}\phantom{.} && \multicolumn{2}{c}{Function $f_3$}\\
\cmidrule{1-2} \cmidrule{4-5}  \cmidrule{7-8} \cmidrule{10-11}
Method&Activation&& $(d,M)=(5,300)$& $(d,M)=(10,500)$ && $(d,M)=(5,500)$ & $(d,M)=(10,1000)$ && $(d,M)=(5,500)$ & $(d,M)=(10,200)$\\
\midrule
ANOVA-RFF&Fourier&&$\num{1.40e-6}$&$\num{1.46e-6}$&&$\num{2.65e-5}$&$\num{2.62e-5}$&&$\num{1.00e-4}$&$\num{9.80e-3}$\\
SHRIMP&Fourier&&$\num{1.83e-6}$&$\num{5.00e-4}$&&$\num{8.20e-3}$&$\num{5.50e-3}$&&$\num{2.00e-4}$&$\num{3.81e-1}$\\
HARFE&Fourier&&$\num{5.82e-1}$&$\num{2.38e-0}$&&$\num{1.38e-1}$&$\num{6.65e-1}$&&$\num{3.64e-0}$&$\num{3.98e-0}$\\
\midrule
kernel ridge reg&&&$\num{5.90e-5}$&$\num{4.40e-4}$&&$\num{7.1e-5}$&$\num{5.10e-4}$&&$\num{1.15e-2}$&$\num{1.69e-0}$\\
\midrule
\multirow{6}{*}{neural net}&\multirow{2}{*}{Fourier}&&$\num{2.36e-4}$&$\num{1.03e-3}$&&$\num{5.28e-5}$&$\num{2.23e-4}$&&$\num{3.14e-3}$&$\num{2.96e-0}$\\
&&&($\lambda=0$)&($\lambda=\num{1e-4}$)&&($\lambda=0$)&($\lambda=\num{1e-4}$)&&($\lambda=\num{1e-4}$)&($\lambda=\num{1e-4}$)\\\cmidrule{2-11}
&\multirow{2}{*}{sigmoid}&&$\num{3.84e-5}$&$\num{5.34e-5}$&&$\num{2.25e-5}$&$\num{3.71e-5}$&&$\num{2.56e-3}$&$\num{2.15e-0}$\\
&&&($\lambda=0$)&($\lambda=\num{1e-4}$)&&($\lambda=\num{1e-4}$)&($\lambda=\num{1e-4}$)&&($\lambda=0$)&($\lambda=\num{1e-3}$)\\\cmidrule{2-11}
&\multirow{2}{*}{ReLU}&&$\num{4.57e-4}$&$\num{1.25e-3}$&&$\num{1.21e-4}$&$\num{1.65e-4}$&&$\num{6.77e-2}$&$\num{1.55e-0}$\\
&&&($\lambda=\num{1e-4}$)&($\lambda=0$)&&($\lambda=0$)&($\lambda=0$)&&($\lambda=\num{1e-4}$)&($\lambda=\num{1e-3}$)\\
\midrule
\multirow{4}{*}{F-Opt}&\multirow{2}{*}{Fourier}&&$\num{2.50e-3}$&$\num{1.08e-0}$&&$\num{2.36e-6}$&$\num{1.31e-0}$&&$\num{5.93e-2}$&$\num{1.68e+1}$\\
&&&($\lambda=\num{1e-4}$)&($\lambda=\num{1e-3}$)&&($\lambda=\num{1e-4}$)&($\lambda=\num{1e-1}$)&&($\lambda=0$)&($\lambda=\num{1e-4}$)\\\cmidrule{2-11}
&\multirow{2}{*}{sigmoid}&&$\num{1.15e-6}$&$\num{2.77e-4}$&&$\num{1.43e-6}$&$\num{9.54e-6}$&&$\num{5.45e-4}$&$\num{2.91e-0}$\\
&&&($\lambda=0$)&($\lambda=0$)&&($\lambda=0$)&($\lambda=0$)&&($\lambda=0$)&($\lambda=0$)\\
\midrule
\multirow{4}{*}{GFT}&\multirow{2}{*}{Fourier}&&$\num{2.72e-7}$&$\num{5.00e-7}$&&$\num{1.03e-7}$&$\num{4.09e-7}$&&$\num{5.87e-5}$&$\num{4.47e-3}$\\
&&&($\lambda=0$)&($\lambda=0$)&&($\lambda=0$)&($\lambda=0$)&&($\lambda=\num{1e-4})$&($\lambda=\num{1e-4}$)\\\cmidrule{2-11}
&\multirow{2}{*}{sigmoid}&&$\num{3.18e-6}$&$\num{1.81e-6}$&&$\num{4.09e-7}$&$\num{6.01e-7}$&&$\num{6.40e-4}$&$\num{1.18e-2}$\\
&&&($\lambda=0$)&($\lambda=0$)&&($\lambda=0$)&($\lambda=0$)&&($\lambda=0$)&($\lambda=\num{1e-4}$)\\
\midrule
\multirow{4}{*}{GFT-r}&\multirow{2}{*}{Fourier}&&\best$\num{6.05e-8}$&\best$\num{5.46e-8}$&&\best$\num{2.02e-8}$&\best$\num{8.15e-8}$&&\best$\num{6.26e-6}$&\best$\num{1.89e-4}$\\
&&&($\lambda=0$)&($\lambda=0$)&&($\lambda=0$)&($\lambda=0$)&&($\lambda=0$)&($\lambda=0$)\\\cmidrule{2-11}
&\multirow{2}{*}{sigmoid}&&$\num{1.05e-6}$&$\num{5.60e-7}$&&$\num{4.97e-8}$&$\num{1.12e-7}$&&$\num{1.50e-5}$&$\num{9.94e-3}$\\
&&&($\lambda=0$)&($\lambda=0$)&&($\lambda=0$)&($\lambda=0$)&&($\lambda=0$)&($\lambda=\num{1e-4}$)\\
\bottomrule
\end{tabular}
}\label{tab:results_functions}
\end{table}

So far, we considered functions $f_i$ that can be represented as sums, where each summand only depends on a small number of inputs $x_i$. 
While this assumption is crucial for the sparse Fourier feature methods from Table~\ref{tab:results_functions}, it is not required for our methods.
Therefore, we also benchmark them on the following non-decomposable functions  and compare the results with standard gradient-based neural network training:
\begin{itemize}
    \item $h_1(x)=\sin(\sum_{i=1}^d x_i)+\|x\|_2^2$
    \item $h_2(x)=\sqrt{\|x-\frac12 e\|_1}$, where $e$ is the vector with all entries equal to one
    \item $h_3(x)=\sqrt{f_3(x)}=\sqrt{10\sin(\pi x_1x_2)+20(x_3-\frac12)^2+10x_4+5x_5}$.
\end{itemize}
The results are given in Table~\ref{tab:results_functions_non_anova}.
As in the previous case, we can see a clear advantage of GFT and GFT-r.

\begin{table}[tbp]
    \centering
    \caption{Function approximation: We report the MSE over the test set averaged over 5 runs.
    The deployed $\lambda$ is indicated below each result. The best performance is highlighted.}
    \vspace{.2cm}
    
   \scalebox{.65}{
\begin{tabular}{ccclllll}
\toprule
\multicolumn{2}{c}{Method} & \phantom{.}   &  \multicolumn{1}{c}{Function $h_1$}  &  \phantom{.} & \multicolumn{1}{c}{Function $h_2$}\phantom{.} && \multicolumn{1}{c}{Function $h_3$}\\
\cmidrule{1-2} \cmidrule{4-4}  \cmidrule{6-6} \cmidrule{8-8}
Method&Activation&&$(d,M)=(10,1000)$ && $(d,M)=(20,1000)$ &&$(d,M)=(5,500)$\\
\midrule
kernel ridge reg&&&$\num{8.91e-2}$&&$\num{3.74e-3}$&&$\num{5.55e-3}$\\\midrule
\multirow{6}{*}{neural net}&\multirow{2}{*}{Fourier}&&$\num{6.03e-2}$&&$\num{1.34e-2}$&&$\num{2.68e-4}$\\
&&&($\lambda=\num{1e-3}$)&&($\lambda=0$)&&($\lambda=0$)\\\cmidrule{2-8}
&\multirow{2}{*}{sigmoid}&&$\num{4.17e-2}$&&$\num{5.94e-3}$&&$\num{4.42e-4}$\\
&&&($\lambda=0$)&&($\lambda=\num{1e-4}$)&&($\lambda=\num{1e-4}$)\\\cmidrule{2-8}
&\multirow{2}{*}{ReLU}&&$\num{5.64e-1}$&&$\num{6.89e-3}$&&$\num{5.56e-3}$\\
&&&($\lambda=\num{1e-4}$)&&($\lambda=\num{1e-4}$)&&($\lambda=\num{1e-3}$)\\
\midrule
\multirow{4}{*}{F-Opt}&\multirow{2}{*}{Fourier}&&$\num{4.27e+1}$&&$\num{5.06e-0}$&&$\num{7.24e-3}$\\
&&&($\lambda=\num{1e-0}$)&&($\lambda=\num{1e-4}$)&&($\lambda=\num{1e-3}$)\\\cmidrule{2-8}
&\multirow{2}{*}{sigmoid}&&$\num{6.43e-2}$&&$\num{7.08e-3}$&&$\num{2.35e-4}$\\
&&&($\lambda=0$)&&($\lambda=0$)&&($\lambda=0$)\\
\midrule
\multirow{4}{*}{GFT}&\multirow{2}{*}{Fourier}&&$\num{2.62e-2}$&&$\num{3.54e-3}$&&$\num{2.34e-4}$\\
&&&($\lambda=\num{1e-3}$)&&($\lambda=\num{1e-4}$)&&($\lambda=\num{1e-4}$)\\\cmidrule{2-8}
&\multirow{2}{*}{sigmoid}&&$\num{9.36e-2}$&&$\num{1.10e-2}$&&$\num{4.70e-4}$\\
&&&($\lambda=\num{1e-4}$)&&($\lambda=\num{1e-4}$)&&($\lambda=\num{1e-4}$)\\
\midrule
\multirow{4}{*}{GFT-r}&\multirow{2}{*}{Fourier}&&\best$\num{8.96e-3}$&&\best $\num{2.57e-3}$&&\best$\num{1.04e-4}$\\
&&&($\lambda=0$)&&($\lambda=\num{1e-4}$)&&($\lambda=0$)\\\cmidrule{2-8}
&\multirow{2}{*}{sigmoid}&&$\num{6.06e-2}$&&$\num{1.00e-2}$&&$\num{2.84e-4}$\\
&&&($\lambda=\num{1e-3}$)&&($\lambda=\num{1e-4}$)&&($\lambda=\num{1e-4}$)\\
\bottomrule
\end{tabular}
}

    \label{tab:results_functions_non_anova}
\end{table}

\subsection{Regression on UCI Datasets}

Next, we apply our method for regression on several UCI datasets \cite{uci}.
For this, we do not have an underlying ground truth function $f$.
Here, we compare our methods with standard gradient-based neural network training, SHRIMP and SALSA.
To this end, we use the numerical setup of SHRIMP.
For each dataset, the MSE on the test split is given in Table~\ref{tab:datasets}.
Compared to the other methods, SHRIMP and SALSA appear a bit more robust to noise and outliers, which frequently occur in the UCI datasets.
This behavior is not surprising, since the enforced sparsity of the features $w_l$ for those methods is a strong implicit regularization.
Incorporating similar sparsity constraints into our generative training is left for future research.
Even without such a regularization, GFT-r  achieves the best performance on most datasets.
Again, both GFT and GFT-r achieve significantly better results than standard training with the Adam optimizer. 

\begin{table}[tbp]
    \centering
    \caption{Regression on UCI datasets: We report the MSE on the test datasets averaged over 5 runs. The values for SHRIMP and SALSA are taken from \citet{DoLiWa2022}.
    The deployed $\lambda$ is indicated below each result. The best performance is highlighted.}
    \vspace{.2cm}
    
   \scalebox{.65}{
\begin{tabular}{cccllllll}
\toprule
\multicolumn{2}{c}{Method}  & \phantom{.}  & \multicolumn{6}{c}{Dataset}\\
\cmidrule{1-2}  \cmidrule{4-9}
\multirow{2}{*}{Method}& \multirow{2}{*}{Activation} & & \multicolumn{1}{c}{Propulsion}&\multicolumn{1}{c}{Galaxy}&\multicolumn{1}{c}{Airfoil}&\multicolumn{1}{c}{CCPP}&\multicolumn{1}{c}{Telemonit}&\multicolumn{1}{c}{Skillkraft}\\
&&&$(d,M)=(15,200)$&$(d,M)=(20,2000)$&$(d,M)=(41,750)$&$(d,M)=(59,2000)$&$(d,M)=(19,1000)$&$(d,M)=(18,1700)$\\\midrule
SHRIMP&Fourier&&$\num{1.02e-6}$&$\num{5.41e-6}$&$\num{2.65e-1}$&\best$\num{6.55e-2}$&$\num{6.00e-2}$&$\num{5.81e-1}$\\
SALSA&Fourier&&$\num{8.81e-3}$&$\num{1.35e-4}$&$\num{5.18e-1}$&$\num{6.78e-2}$&$\num{3.47e-2}$&\best$\num{5.47e-1}$\\\midrule
kernel ridge reg&&&$\num{8.60e-3}$&$\num{2.38e-3}$&$\num{8.10e-1}$&$\num{1.24e-1}$&$\num{1.06e-1}$&$\num{6.30e-0}$\\\midrule
\multirow{6}{*}{neural net}&\multirow{2}{*}{Fourier}&&$\num{9.07e-3}$&$\num{4.46e-4}$&$\num{3.41e-1}$&$\num{6.97e-2}$&$\num{2.51e-2}$&$\num{6.01e-1}$\\
&&&($\lambda=\num{1e-2}$)&($\lambda=\num{1e-4}$)&($\lambda=\num{1e-1}$)&($\lambda=\num{1e-1}$)&($\lambda=\num{1e-3}$)&($\lambda=\num{1e-1}$)\\\cmidrule{2-9}
&\multirow{2}{*}{sigmoid}&&$\num{9.21e-3}$&$\num{1.67e-4}$&$\num{3.31e-1}$&$\num{8.01e-2}$&$\num{7.86e-2}$&$\num{1.57e-0}$\\
&&&($\lambda=0$)&($\lambda=\num{1e-4}$)&($\lambda=\num{1e-1}$)&($\lambda=\num{1e-1}$)&($\lambda=\num{1e-3}$)&($\lambda=\num{1e-3}$)\\\cmidrule{2-9}
&\multirow{2}{*}{ReLU}&&$\num{5.92e-4}$&$\num{4.72e-4}$&$\num{3.66e-1}$&$\num{6.73e-2}$&$\num{2.71e-2}$&$\num{2.23e-0}$\\
&&&($\lambda=\num{1e-3}$)&($\lambda=0$)&($\lambda=\num{1e-1}$)&($\lambda=\num{1e-1}$)&($\lambda=\num{1e-2}$)&($\lambda=\num{1e-0}$)\\
\midrule
\multirow{4}{*}{F-Opt}&\multirow{2}{*}{Fourier}&&$\num{6.96e-1}$&$\num{3.51e0}$&$\num{1.05e0}$&$\num{9.97e-1}$&$\num{1.01e-0}$&$\num{1.01e-0}$\\
&&&($\lambda=\num{1e-3}$)&($\lambda=\num{1e-1}$)&($\lambda=\num{1e-4}$)&($\lambda=\num{1e-1}$)&($\lambda=\num{1e-1}$)&($\lambda=\num{1e-0}$)\\\cmidrule{2-9}
&\multirow{2}{*}{sigmoid}&&$\num{1.57e-2}$&$\num{1.91e-4}$&$\num{5.92e-1}$&$\num{7.35e-2}$&$\num{7.38e-2}$&$\num{5.79e-1}$\\
&&&($\lambda=\num{1e-4}$)&($\lambda=0$)&($\lambda=\num{1e-3}$)&($\lambda=\num{1e-1}$)&($\lambda=\num{1e-3}$)&($\lambda=\num{1e-0}$)\\
\midrule
\multirow{4}{*}{GFT}&\multirow{2}{*}{Fourier}&&$\num{8.31e-7}$&$\num{3.31e-5}$&\best$\num{2.34e-1}$&$\num{8.06e-2}$&$\num{1.05e-2}$&$\num{5.66e-1}$\\
&&&($\lambda=0$)&($\lambda=0$)&($\lambda=\num{1e-1}$)&($\lambda=\num{1e-2}$)&($\lambda=\num{1e-2}$)&($\lambda=\num{1e-0}$)\\\cmidrule{2-9}
&\multirow{2}{*}{sigmoid}&&$\num{1.22e-5}$&$\num{7.42e-5}$&$\num{2.90e-1}$&$\num{6.86e-2}$&$\num{1.35e-2}$&$\num{9.68e-1}$\\
&&&($\lambda=0$)&($\lambda=0$)&($\lambda=\num{1e-1}$)&($\lambda=\num{1e-1}$)&($\lambda=\num{1e-4}$)&($\lambda=\num{1e-1}$)\\\midrule
\multirow{4}{*}{GFT-r}&\multirow{2}{*}{Fourier}&&\best$\num{6.97e-7}$&\best$\num{5.36e-6}$&\best$\num{2.34e-1}$&$\num{8.04e-2}$&\best$\num{6.48e-3}$&$\num{5.65e-1}$\\
&&&($\lambda=0$)&($\lambda=0$)&($\lambda=\num{1e-1}$)&($\lambda=\num{1e-2}$)&($\lambda=\num{0}$)&($\lambda=\num{1e-0}$)\\\cmidrule{2-9}
&\multirow{2}{*}{sigmoid}&&$\num{1.67e-5}$&$\num{1.85e-5}$&$\num{2.89e-1}$&$\num{6.84e-2}$&$\num{9.39e-3}$&$\num{9.88e-1}$\\
&&&($\lambda=0$)&($\lambda=0$)&($\lambda=\num{1e-1}$)&($\lambda=\num{1e-1}$)&($\lambda=0$)&($\lambda=\num{1e-1}$)\\
\bottomrule
\end{tabular}
}

    \label{tab:datasets}
\end{table}

\section{Discussion}

\paragraph{Summary}
We proposed a training procedure for $f_{w,b}$ as in \eqref{eq:Architecture}  with only a few hidden neurons $w$.
In our procedure, we sample the $w$ from a generative model and compute the optimal $b$ by solving a linear system.
To enhance the results, we apply a feature refinement scheme in the latent space of the generative model and regularize the loss function.
Numerical examples have shown that the proposed generative feature training significantly outperforms standard training procedures.

\paragraph{Outlook}
Our approach can be extended in several directions.
First, we want to train deeper networks in a greedy way similar to \citep{BelEicOya2019}.
Recently, a similar approach was considered in the context of sampled networks by \citet{BolBurDat2023}.
Moreover, we can encode a sparse structure on the features by replacing the latent distribution $N(0,I_d)$ with a lower-dimensional latent model or by considering mixtures of generative models.
From a theoretical side, we want to characterize the global minimizers of the functional in \eqref{eq:loss_von_theta} and their relations to the Fourier transform of the target function. 

\paragraph{Limitations}
If $N$ in \eqref{eq:Architecture} gets large, solving the linear system \eqref{eq:b_von_w} becomes expensive.
However, this corresponds to the overparameterized regime where gradient-based methods work well.
Moreover, the computation of the optimal $b$ depends on all data points.
Consequently, if we do minibatching, the output weights $b$ are batch-dependent, and both the theoretical and practical implications remain open.
Instead, we emphasize that one motivation for our method is the treatment of small data sets, where no minibatching is required.
This is actually also one of the main use cases for 2-layer neural networks. Finally, note that GFT is currently restricted to the $L_2$-loss function, which limits the applicability of GFT to non-regression tasks. For other loss functions, bilevel learning methods could be used to compute and differentiate the optimal output layer.
This is beyond the scope of this paper, and we leave this point for future work.

\subsubsection*{Acknowledgments}
We would like to thank Daniel Potts, Gabriele Steidl and Laura Weidensager for fruitful discussions.
JH acknwoledges funding by the German Research Foundation (DFG) within the Walter Benjamin Programme with project number 530824055 and by the EPSRC programme grant ``The Mathematics of Deep Learning'' with reference EP/V026259/1.

\bibliography{ref}

\begin{thebibliography}{45}
\providecommand{\natexlab}[1]{#1}
\providecommand{\url}[1]{\texttt{#1}}
\expandafter\ifx\csname urlstyle\endcsname\relax
  \providecommand{\doi}[1]{doi: #1}\else
  \providecommand{\doi}{doi: \begingroup \urlstyle{rm}\Url}\fi

\bibitem[Acar \& Vogel(1994)Acar and Vogel]{AcaVog1994}
Robert Acar and Curtis~R Vogel.
\newblock Analysis of bounded variation penalty methods for ill-posed problems.
\newblock \emph{Inverse Problems}, 10\penalty0 (6):\penalty0 1217--1229, 1994.

\bibitem[Ambrosio et~al.(2000)Ambrosio, Fusco, and Pallara]{Ambrosio2000}
Luigi Ambrosio, Nicola Fusco, and Diego Pallara.
\newblock \emph{Functions of Bounded Variation and Free Discontinuity
  Problems}.
\newblock Oxford Mathematical Monographs. Oxford University Press, New York,
  2000.

\bibitem[Ba et~al.(2024)Ba, Melnyk, Wald, and Steidl]{BMWS2024}
Fatima~Antarou Ba, Oleh Melnyk, Christian Wald, and Gabriele Steidl.
\newblock Sparse additive function decompositions facing basis transforms.
\newblock \emph{Foundations of Data Science}, 6\penalty0 (4):\penalty0
  514--552, 2024.

\bibitem[Bai et~al.(2023)Bai, Gautam, and Sojoudi]{BaiGauSoj2023}
Yatong Bai, Tanmay Gautam, and Somayeh Sojoudi.
\newblock Efficient global optimization of two-layer {ReLU} networks:
  Quadratic-time algorithms and adversarial training.
\newblock \emph{SIAM Journal on Mathematics of Data Science}, 5\penalty0
  (2):\penalty0 446--474, 2023.

\bibitem[Bai et~al.(2024)Bai, Lu, and Zhang]{BaiLuZha2024}
Yaxuan Bai, Xiaofan Lu, and Linan Zhang.
\newblock Function approximations via $\ell_1$-$\ell_2$ optimization.
\newblock \emph{Journal of Applied \& Numerical Optimization}, 6\penalty0
  (3):\penalty0 371--389, 2024.

\bibitem[Barbu(2023)]{Barbu2023}
Adrian Barbu.
\newblock Training a two-layer {ReLU} network analytically.
\newblock \emph{Sensors}, 23\penalty0 (8):\penalty0 4072, 2023.

\bibitem[Belilovsky et~al.(2019)Belilovsky, Eickenberg, and
  Oyallon]{BelEicOya2019}
Eugene Belilovsky, Michael Eickenberg, and Edouard Oyallon.
\newblock Greedy layerwise learning can scale to {I}magenet.
\newblock In \emph{International Conference on Machine Learning}, pp.\
  583--593. PMLR, 2019.

\bibitem[Blundell et~al.(2015)Blundell, Cornebise, Kavukcuoglu, and
  Wierstra]{BluCorKav2015}
Charles Blundell, Julien Cornebise, Koray Kavukcuoglu, and Daan Wierstra.
\newblock Weight uncertainty in neural network.
\newblock In \emph{International Conference on Machine Learning}, pp.\
  1613--1622. PMLR, 2015.

\bibitem[Bolager et~al.(2023)Bolager, Burak, Datar, Sun, and
  Dietrich]{BolBurDat2023}
Erik~Lien Bolager, Iryna Burak, Chinmay Datar, Qing Sun, and Felix Dietrich.
\newblock Sampling weights of deep neural networks.
\newblock In \emph{Advances in Neural Information Processing Systems},
  volume~37, 2023.

\bibitem[Boob et~al.(2022)Boob, Dey, and Lan]{BooDey2022}
Digvijay Boob, Santanu~S Dey, and Guanghui Lan.
\newblock Complexity of training {ReLU} neural network.
\newblock \emph{Discrete Optimization}, 44:\penalty0 100620, 2022.

\bibitem[Chan \& Esedoglu(2005)Chan and Esedoglu]{ChanEse2005}
Tony~F Chan and Selim Esedoglu.
\newblock Aspects of total variation regularized $l^1$ function approximation.
\newblock \emph{SIAM Journal on Applied Mathematics}, 65\penalty0 (5):\penalty0
  1817--1837, 2005.

\bibitem[Cortes et~al.(2010)Cortes, Mohri, and Talwalkar]{CorMohTal2010}
Corinna Cortes, Mehryar Mohri, and Ameet Talwalkar.
\newblock On the impact of kernel approximation on learning accuracy.
\newblock In \emph{International Conference on Artificial Intelligence and
  Statistics}, pp.\  113--120, 2010.

\bibitem[Cristianini \& Shawe-Taylor(2000)Cristianini and Shawe-Taylor]{CS2000}
Nello Cristianini and John Shawe-Taylor.
\newblock \emph{An introduction to support vector machines and other
  kernel-based learning methods}.
\newblock Cambridge University Press, 2000.

\bibitem[Dunbar et~al.(2025)Dunbar, Nelsen, and Mutic]{DNM2025}
Oliver~RA Dunbar, Nicholas~H Nelsen, and Maya Mutic.
\newblock Hyperparameter optimization for randomized algorithms: a case study
  on random features.
\newblock \emph{Statistics and Computing}, 35\penalty0 (3):\penalty0 1--28,
  2025.

\bibitem[Dutt \& Rokhlin(1993)Dutt and Rokhlin]{DR1993}
Alok Dutt and Vladimir Rokhlin.
\newblock Fast {F}ourier transforms for nonequispaced data.
\newblock \emph{SIAM Journal on Scientific Computing}, 14\penalty0
  (6):\penalty0 1368--1393, 1993.

\bibitem[E et~al.(2019)E, Ma, and Wu]{EMaWu2019}
Weinan E, Chao Ma, and Lei Wu.
\newblock A priori estimates of the population risk for two-layer neural
  networks.
\newblock \emph{Communications in Mathematical Sciences}, 17\penalty0
  (5):\penalty0 1407--1425, 2019.

\bibitem[Graves(2011)]{Graves2011}
Alex Graves.
\newblock Practical variational inference for neural networks.
\newblock In \emph{Advances in Neural Information Processing Systems},
  volume~24, 2011.

\bibitem[Hashemi et~al.(2023)Hashemi, Schaeffer, Shi, Topcu, Tran, and
  Ward]{HasSchShi2023}
Abolfazl Hashemi, Hayden Schaeffer, Robert Shi, Ufuk Topcu, Giang Tran, and
  Rachel Ward.
\newblock Generalization bounds for sparse random feature expansions.
\newblock \emph{Applied and Computational Harmonic Analysis}, 62:\penalty0
  310--330, 2023.

\bibitem[Hertrich(2024)]{H2024}
Johannes Hertrich.
\newblock Fast kernel summation in high dimensions via slicing and {F}ourier
  transforms.
\newblock \emph{SIAM Journal on Mathematics of Data Science}, 6:\penalty0
  1109--1137, 2024.

\bibitem[Hertrich et~al.(2025)Hertrich, Jahn, and Quellmalz]{HJQ2024}
Johannes Hertrich, Tim Jahn, and Michael Quellmalz.
\newblock Fast summation of radial kernels via {QMC} slicing.
\newblock \emph{International Conference on Learning Representations}, 2025.

\bibitem[Holzm{\"u}ller \& Steinwart(2022)Holzm{\"u}ller and
  Steinwart]{HolSte2022}
David Holzm{\"u}ller and Ingo Steinwart.
\newblock Training two-layer {ReLU} networks with gradient descent is
  inconsistent.
\newblock \emph{Journal of Machine Learning Research}, 23\penalty0
  (181):\penalty0 1--82, 2022.

\bibitem[Huang et~al.(2006)Huang, Zhu, and Siew]{HZS2006}
Guang-Bin Huang, Qin-Yu Zhu, and Chee-Kheong Siew.
\newblock Extreme learning machine: theory and applications.
\newblock \emph{Neurocomputing}, 70\penalty0 (1-3):\penalty0 489--501, 2006.

\bibitem[Jospin et~al.(2022)Jospin, Laga, Boussaid, Buntine, and
  Bennamoun]{JosLagBou2022}
Laurent~Valentin Jospin, Hamid Laga, Farid Boussaid, Wray Buntine, and Mohammed
  Bennamoun.
\newblock Hands-on {B}ayesian neural networks—a tutorial for deep learning
  users.
\newblock \emph{IEEE Computational Intelligence Magazine}, 17\penalty0
  (2):\penalty0 29--48, 2022.

\bibitem[Kammonen et~al.(2020)Kammonen, Kiessling, Plech{\'a}{\v{c}}, Sandberg,
  and Szepessy]{KamKiePle2020}
Aku Kammonen, Jonas Kiessling, Petr Plech{\'a}{\v{c}}, Mattias Sandberg, and
  Anders Szepessy.
\newblock Adaptive random {F}ourier features with {M}etropolis sampling.
\newblock \emph{Foundations of Data Science}, 2\penalty0 (3):\penalty0
  309--332, 2020.

\bibitem[Kandasamy \& Yu(2016)Kandasamy and Yu]{KanYu2016}
Kirthevasan Kandasamy and Yaoliang Yu.
\newblock Additive approximations in high dimensional nonparametric regression
  via the {SALSA}.
\newblock In \emph{International Conference on Machine Learning}, pp.\  69--78.
  PMLR, 2016.

\bibitem[Kelly et~al.(2023)Kelly, Longjohn, and Nottingham]{uci}
Markelle Kelly, Rachel Longjohn, and Kolby Nottingham.
\newblock The {UCI} machine learning repository, 2023.
\newblock URL \url{https://archive.ics.uci.edu}.

\bibitem[Li et~al.(2019{\natexlab{a}})Li, Chang, Mroueh, Yang, and
  Poczos]{LCMYP2019}
Chun-Liang Li, Wei-Cheng Chang, Youssef Mroueh, Yiming Yang, and Barnabas
  Poczos.
\newblock Implicit kernel learning.
\newblock In \emph{International Conference on Artificial Intelligence and
  Statistics}, pp.\  2007--2016. PMLR, 2019{\natexlab{a}}.

\bibitem[Li et~al.(2022)Li, Tai, and Yang]{LiXueYan2022}
Lingfeng Li, Xue-Cheng Tai, and Jiang Yang.
\newblock Generalization error analysis of neural networks with gradient based
  regularization.
\newblock \emph{Communications in Computational Physics}, 32\penalty0
  (4):\penalty0 1007--1038, 2022.

\bibitem[Li et~al.(2021)Li, Si, Li, Hsieh, and Bengio]{LiSiLi2021}
Yang Li, Si~Si, Gang Li, Cho-Jui Hsieh, and Samy Bengio.
\newblock Learnable {F}ourier features for multi-dimensional spatial positional
  encoding.
\newblock \emph{Advances in Neural Information Processing Systems},
  34:\penalty0 15816--15829, 2021.

\bibitem[Li et~al.(2019{\natexlab{b}})Li, Zhang, Wang, and Kumar]{LiZhaWan2019}
Yanjun Li, Kai Zhang, Jun Wang, and Sanjiv Kumar.
\newblock Learning adaptive random features.
\newblock In \emph{Proceedings of the AAAI Conference on Artificial
  Intelligence}, volume~33, pp.\  4229--4236, 2019{\natexlab{b}}.

\bibitem[Li et~al.(2019{\natexlab{c}})Li, Ton, Oglic, and
  Sejdinovic]{LiTonOgl2019}
Zhu Li, Jean-Francois Ton, Dino Oglic, and Dino Sejdinovic.
\newblock Towards a unified analysis of random {F}ourier features.
\newblock In \emph{International Conference on Machine Learning}, pp.\
  3905--3914. PMLR, 2019{\natexlab{c}}.

\bibitem[Liu et~al.(2021)Liu, Huang, Chen, and Suykens]{LiuHuaChe2021}
Fanghui Liu, Xiaolin Huang, Yudong Chen, and Johan~AK Suykens.
\newblock Random features for kernel approximation: A survey on algorithms,
  theory, and beyond.
\newblock \emph{IEEE Transactions on Pattern Analysis and Machine
  Intelligence}, 44\penalty0 (10):\penalty0 7128--7148, 2021.

\bibitem[Mishkin et~al.(2022)Mishkin, Sahiner, and Pilanci]{MisSahPil2022}
Aaron Mishkin, Arda Sahiner, and Mert Pilanci.
\newblock Fast convex optimization for two-layer {ReLU} networks: Equivalent
  model classes and cone decompositions.
\newblock In \emph{International Conference on Machine Learning}, pp.\
  15770--15816. PMLR, 2022.

\bibitem[Neal(2012)]{Neal2012}
Radford~M Neal.
\newblock \emph{Bayesian Learning for Neural Networks}.
\newblock Springer Science \& Business Media, 2012.

\bibitem[Pilanci \& Ergen(2020)Pilanci and Ergen]{PilErg2020}
Mert Pilanci and Tolga Ergen.
\newblock Neural networks are convex regularizers: Exact polynomial-time convex
  optimization formulations for two-layer networks.
\newblock In \emph{International Conference on Machine Learning}, pp.\
  7695--7705. PMLR, 2020.

\bibitem[Potts \& Schmischke(2021)Potts and Schmischke]{PotSch2021}
Daniel Potts and Michael Schmischke.
\newblock Interpretable approximation of high-dimensional data.
\newblock \emph{SIAM Journal on Mathematics of Data Science}, 3\penalty0
  (4):\penalty0 1301--1323, 2021.

\bibitem[Potts \& Weidensager(2025)Potts and Weidensager]{PW2024}
Daniel Potts and Laura Weidensager.
\newblock {ANOVA}-boosting for random {F}ourier features.
\newblock \emph{Applied and Computational Harmonic Analysis}, 79:\penalty0
  101789, 2025.

\bibitem[Potts et~al.(2001)Potts, Steidl, and Tasche]{PST2001}
Daniel Potts, Gabriele Steidl, and Manfred Tasche.
\newblock Fast {F}ourier transforms for nonequispaced data: A tutorial.
\newblock \emph{Modern Sampling Theory: Mathematics and Applications}, pp.\
  247--270, 2001.

\bibitem[Rahimi \& Recht(2007)Rahimi and Recht]{RahRec2007}
Ali Rahimi and Benjamin Recht.
\newblock Random features for large-scale kernel machines.
\newblock In \emph{Advances in Neural Information Processing Systems},
  volume~20, 2007.

\bibitem[Rahimi \& Recht(2008)Rahimi and Recht]{RahRec2008}
Ali Rahimi and Benjamin Recht.
\newblock Uniform approximation of functions with random bases.
\newblock In \emph{46th Annual Allerton Conference on Communication, Control,
  and Computing}, pp.\  555--561. IEEE, 2008.

\bibitem[Rudi \& Rosasco(2017)Rudi and Rosasco]{RudRos2017}
Alessandro Rudi and Lorenzo Rosasco.
\newblock Generalization properties of learning with random features.
\newblock In \emph{Advances in Neural Information Processing Systems},
  volume~30, 2017.

\bibitem[Rux et~al.(2025)Rux, Quellmalz, and Steidl]{RQS2024}
Nicolaj Rux, Michael Quellmalz, and Gabriele Steidl.
\newblock Slicing of radial functions: a dimension walk in the {F}ourier space.
\newblock \emph{Sampling Theory, Signal Processing, and Data Analysis},
  23\penalty0 (1):\penalty0 1--40, 2025.

\bibitem[Saha et~al.(2023)Saha, Schaeffer, and Tran]{SaScTr2023}
Esha Saha, Hayden Schaeffer, and Giang Tran.
\newblock {HARFE}: Hard-ridge random feature expansion.
\newblock \emph{Sampling Theory, Signal Processing, and Data Analysis},
  21\penalty0 (2):\penalty0 27, 2023.

\bibitem[Xie et~al.(2022)Xie, Shi, Schaeffer, and Ward]{DoLiWa2022}
Yuege Xie, Robert Shi, Hayden Schaeffer, and Rachel Ward.
\newblock {SHRIMP}: Sparser random feature models via iterative magnitude
  pruning.
\newblock In \emph{Proceedings of Mathematical and Scientific Machine
  Learning}, volume 190, pp.\  303--318. PMLR, 2022.

\bibitem[Yen et~al.(2014)Yen, Lin, Lin, Ravikumar, and Dhillon]{YenLinLin2014}
Ian En-Hsu Yen, Ting-Wei Lin, Shou-De Lin, Pradeep~K Ravikumar, and Inderjit~S
  Dhillon.
\newblock Sparse random feature algorithm as coordinate descent in {H}ilbert
  space.
\newblock In \emph{Advances in Neural Information Processing Systems},
  volume~27, 2014.

\end{thebibliography}
\bibliographystyle{tmlr}

\appendix

\section{Dependence on the Number of Features}\label{app:other_N}

We redo the experiments from Section~\ref{subsec:function_approx} for $N=50$ and $N=200$. The results are given in Table~\ref{tab:results_functions_50} and \ref{tab:results_functions_200}.

\begin{table}[tbp]
    \centering
    \caption{Comparison for function approximation with: We report the MSE over the test set averaged over 5 runs.
    The table contains the same experiments as Table~\ref{tab:results_functions} with $N=50$ features.
    The deployed $\lambda$ is indicated below each result. The best performance is highlighted.}
    \vspace{.2cm}
    
   \scalebox{.65}{
\begin{tabular}{cccllllllll}
\toprule
\multicolumn{2}{c}{Method} & \phantom{.}   &  \multicolumn{2}{c}{Function $f_1$}  &  \phantom{.} & \multicolumn{2}{c}{Function $f_2$}\phantom{.} && \multicolumn{2}{c}{Function $f_3$}\\
\cmidrule{1-2} \cmidrule{4-5}  \cmidrule{7-8} \cmidrule{10-11}
Method&Activation&& $(d,M)=(5,300)$& $(d,M)=(10,500)$ && $(d,M)=(5,500)$ & $(d,M)=(10,1000)$ && $(d,M)=(5,500)$ & $(d,M)=(10,200)$\\
\midrule
\multirow{6}{*}{neural net}&\multirow{2}{*}{Fourier}&&$\num{3.35e-5}$&$\num{9.22e-5}$&&$\num{6.97e-6}$&$\num{8.15e-6}$&&$\num{2.45e-3}$&$\num{4.65e-0}$\\
&&&($\lambda=0$)&($\lambda=\num{1e-4}$)&&($\lambda=0$)&($\lambda=0$)&&($\lambda=\num{1e-4}$)&($\lambda=\num{1e-3}$)\\\cmidrule{2-11}
&\multirow{2}{*}{sigmoid}&&$\num{1.03e-5}$&$\num{1.52e-5}$&&$\num{1.19e-5}$&$\num{1.45e-5}$&&$\best\num{2.14e-3}$&$\num{2.27e-0}$\\
&&&($\lambda=0$)&($\lambda=0$)&&($\lambda=0$)&($\lambda=\num{1e-4}$)&&($\lambda=0$)&($\lambda=\num{1e-4}$)\\\cmidrule{2-11}
&\multirow{2}{*}{ReLU}&&$\num{4.71e-4}$&$\num{1.40e-3}$&&$\num{3.00e-4}$&$\num{1.12e-4}$&&$\num{2.08e-1}$&$\num{1.89e-0}$\\
&&&($\lambda=0$)&($\lambda=0$)&&($\lambda=0$)&($\lambda=\num{1e-4}$)&&($\lambda=0$)&($\lambda=\num{1e-4}$)\\
\midrule
\multirow{4}{*}{F-Opt}&\multirow{2}{*}{Fourier}&&$\num{1.60e-5}$&$\num{2.52e-0}$&&$\num{5.01e-6}$&$\num{7.45e-0}$&&$\num{5.13e-3}$&$\num{2.60e+2}$\\
&&&($\lambda=\num{1e-4}$)&($\lambda=0$)&&($\lambda=0$)&($\lambda=0$)&&($\lambda=\num{1e-4}$)&($\lambda=\num{1e-4}$)\\\cmidrule{2-11}
&\multirow{2}{*}{sigmoid}&&$\num{2.44e-4}$&$\num{8.05e-5}$&&$\num{3.58e-3}$&$\num{1.69e-6}$&&$\num{1.08e-0}$&$\num{3.94e-0}$\\
&&&($\lambda=0$)&($\lambda=\num{1e-4}$)&&($\lambda=\num{1e-4}$)&($\lambda=0$)&&($\lambda=\num{1e-4}$)&($\lambda=\num{1e-4}$)\\
\midrule
\multirow{4}{*}{GFT}&\multirow{2}{*}{Fourier}&&$\num{9.49e-5}$&$\num{4.19e-5}$&&$\num{5.71e-5}$&$\num{1.98e-5}$&&$\num{9.17e-2}$&$\num{8.72e-3}$\\
&&&($\lambda=0$)&($\lambda=\num{1e-3}$)&&($\lambda=\num{1e-3}$)&($\lambda=\num{1e-4}$)&&($\lambda=\num{1e-3})$&($\lambda=\num{1e-4}$)\\\cmidrule{2-11}
&\multirow{2}{*}{sigmoid}&&$\num{1.94e-1}$&$\num{6.08e-5}$&&$\num{5.41e-1}$&$\num{9.01e-4}$&&$\num{9.81e-0}$&$\num{1.09e-2}$\\
&&&($\lambda=\num{1e-4}$)&($\lambda=\num{1e-3}$)&&($\lambda=\num{1e-4}$)&($\lambda=\num{1e-2}$)&&($\lambda=\num{1e-4}$)&($\lambda=0$)\\
\midrule
\multirow{4}{*}{GFT-r}&\multirow{2}{*}{Fourier}&&\best$\num{5.52e-6}$&\best$\num{2.94e-7}$&&\best$\num{1.24e-6}$&\best$\num{9.75e-7}$&&$\num{8.72e-3}$&\best$\num{2.02e-4}$\\
&&&($\lambda=0$)&($\lambda=0$)&&($\lambda=0$)&($\lambda=\num{1e-4}$)&&($\lambda=0$)&($\lambda=0$)\\\cmidrule{2-11}
&\multirow{2}{*}{sigmoid}&&$\num{1.94e-1}$&$\num{6.96e-2}$&&$\num{4.33e-1}$&$\num{4.78e-1}$&&$\num{2.66e0}$&$\num{2.82e-3}$\\
&&&($\lambda=\num{1e-4}$)&($\lambda=0$)&&($\lambda=0$)&($\lambda=\num{1e-4}$)&&($\lambda=0$)&($\lambda=0$)\\
\bottomrule
\end{tabular}
}
    \label{tab:results_functions_50}
\end{table}

\begin{table}[tbp]
    \centering
    \caption{Comparison for function approximation with: We report the MSE over the test set averaged over 5 runs.
    The table contains the same experiments as Table~\ref{tab:results_functions} with $N=200$ features.
    The deployed $\lambda$ is indicated below each result. The best performance is highlighted.}
    \vspace{.2cm}
    
   \scalebox{.65}{
\begin{tabular}{cccllllllll}
\toprule
\multicolumn{2}{c}{Method} & \phantom{.}   &  \multicolumn{2}{c}{Function $f_1$}  &  \phantom{.} & \multicolumn{2}{c}{Function $f_2$}\phantom{.} && \multicolumn{2}{c}{Function $f_3$}\\
\cmidrule{1-2} \cmidrule{4-5}  \cmidrule{7-8} \cmidrule{10-11}
Method&Activation&& $(d,M)=(5,300)$& $(d,M)=(10,500)$ && $(d,M)=(5,500)$ & $(d,M)=(10,1000)$ && $(d,M)=(5,500)$ & $(d,M)=(10,200)$\\
\midrule
\multirow{6}{*}{neural net}&\multirow{2}{*}{Fourier}&&$\num{7.22e-4}$&$\num{1.21e-2}$&&$\num{1.56e-4}$&$\num{1.36e-4}$&&$\num{2.26e-3}$&$\num{3.95e-0}$\\
&&&($\lambda=0$)&($\lambda=\num{1e-4}$)&&($\lambda=\num{1e-4}$)&($\lambda=\num{1e-4}$)&&($\lambda=\num{1e-4}$)&($\lambda=0$)\\\cmidrule{2-11}
&\multirow{2}{*}{sigmoid}&&$\num{1.54e-5}$&$\num{2.35e-5}$&&$\num{1.46e-5}$&$\num{4.91e-5}$&&$\num{1.21e-3}$&$\num{1.79e-0}$\\
&&&($\lambda=0$)&($\lambda=0$)&&($\lambda=\num{1e-4}$)&($\lambda=0$)&&($\lambda=\num{1e-4}$)&($\lambda=0$)\\\cmidrule{2-11}
&\multirow{2}{*}{ReLU}&&$\num{3.03e-4}$&$\num{1.05e-3}$&&$\num{1.78e-4}$&$\num{2.02e-4}$&&$\num{6.20e-2}$&$\num{1.84e-0}$\\
&&&($\lambda=0$)&($\lambda=0$)&&($\lambda=0$)&($\lambda=\num{1e-4}$)&&($\lambda=\num{1e-4}$)&($\lambda=\num{1e-4}$)\\
\midrule
\multirow{4}{*}{F-Opt}&\multirow{2}{*}{Fourier}&&$\num{9.45e-3}$&$\num{1.52e-2}$&&$\num{4.47e-5}$&$\num{2.34e-2}$&&$\num{1.29e-0}$&$\num{2.11e+1}$\\
&&&($\lambda=\num{1e-3}$)&($\lambda=\num{1e-2}$)&&($\lambda=\num{1e-4}$)&($\lambda=\num{1e-2}$)&&($\lambda=\num{1e-2}$)&($\lambda=\num{1e-0}$)\\\cmidrule{2-11}
&\multirow{2}{*}{sigmoid}&&$\num{2.32e-6}$&$\num{5.73e-4}$&&$\num{1.19e-6}$&$\num{8.92e-5}$&&$\num{7.23e-4}$&$\num{2.40e-0}$\\
&&&($\lambda=0$)&($\lambda=\num{1e-4}$)&&($\lambda=0$)&($\lambda=0$)&&($\lambda=0$)&($\lambda=0$)\\
\midrule
\multirow{4}{*}{GFT}&\multirow{2}{*}{Fourier}&&$\num{6.66e-7}$&$\num{2.33e-7}$&&$\num{5.73e-8}$&$\num{6.53e-7}$&&$\num{2.26e-5}$&$\num{1.51e-0}$\\
&&&($\lambda=0$)&($\lambda=0$)&&($\lambda=0$)&($\lambda=0$)&&($\lambda=0)$&($\lambda=\num{1e-4}$)\\\cmidrule{2-11}
&\multirow{2}{*}{sigmoid}&&$\num{2.84e-6}$&$\num{3.52e-6}$&&$\num{2.04e-7}$&$\num{3.05e-7}$&&$\num{1.62e-4}$&$\num{1.75e-2}$\\
&&&($\lambda=0$)&($\lambda=0$)&&($\lambda=0$)&($\lambda=0$)&&($\lambda=0$)&($\lambda=0$)\\
\midrule
\multirow{4}{*}{GFT-r}&\multirow{2}{*}{Fourier}&&\best$\num{1.57e-7}$&\best$\num{6.01e-8}$&&\best$\num{1.42e-8}$&\best$\num{1.66e-7}$&&\best$\num{1.28e-6}$&$\num{1.82e0}$\\
&&&($\lambda=0$)&($\lambda=0$)&&($\lambda=0$)&($\lambda=0$)&&($\lambda=0$)&($\lambda=\num{1e-4}$)\\\cmidrule{2-11}
&\multirow{2}{*}{sigmoid}&&$\num{2.04e-6}$&$\num{2.26e-6}$&&$\num{4.27e-8}$&$\num{5.41e-8}$&&$\num{1.84e-5}$&\best$\num{1.60e-2}$\\
&&&($\lambda=0$)&($\lambda=0$)&&($\lambda=0$)&($\lambda=0$)&&($\lambda=0$)&($\lambda=0$)\\
\bottomrule
\end{tabular}
}
    \label{tab:results_functions_200}
\end{table}

\section{Further Ablations}\label{app:ablations}
First, to demonstrate that using the optimal output weights $b(w)$ from Section~\ref{subsec:optimal_weights} is necessary for GFT, we run the following experiment:
Instead of sampling only features $w_l$ from a generative model $p_w={G_\theta}_\#\mathcal \eta$ with $G_\theta\colon\R^d\to\R^d$ and then computing $b(w)$, we sample pairs $(w_l,b_l)$ from $p_{w,b}=\tilde {G_\theta}_\#\tilde \eta$ for $\tilde G_\theta\colon\R^{d+2}\to\R^d\times\C$ and the $d+2$ dimensional standard normal distribution $\tilde \eta$.
We use the Fourier activation and the same setup as for Table~\ref{tab:results_functions}.
The results in the first row of Table~\ref{tab:results_ablations} are significantly worse than for GFT and GFT-r in Table~\ref{tab:results_functions}. This is not surprising since now each output weight $b_l$ only depends on its corresponding feature $w_l$ instead of all $(w_l)_{l=1}^N$ as for the optimal $b(w)$. 
In particular, $b(w)$ cannot be learned by a gradient-based algorithms since in each update the $w=(w_l)_{l=1}^N$ are resampled.
Thus, there is no persisting correspondence of sampled features with specific output weights.

Second, we test whether the neural network line in Table~\ref{tab:results_functions} can be improved by introducing gradient noise.
This technique can help to escape local minima.
To this end, we optimize the same loss function as for the 2-layer neural network in Table~\ref{tab:results_functions}, but use the noisy stochastic gradient descent $\theta_{n+1}=\theta_n-\rho \nabla L(\theta)+\alpha Z$ with random $Z\sim\mathcal N(0,I)$, step size $\rho$, and noise strength $\alpha=0.1\rho$.
The  obtained results in the second row of Table~\ref{tab:results_ablations} do not improve upon the baseline from Table~\ref{tab:results_functions}. 

\begin{table}[tbp]
    \centering
    \caption{Comparison for function approximation with: We report the MSE over the test set averaged over 5 runs.
    The table contains the same experiments as Table~\ref{tab:results_functions} for the two ablations in Appendix~\ref{app:ablations}.
    The deployed $\lambda$ is indicated below each result.}
    \vspace{.2cm}
    
   \scalebox{.65}{
\begin{tabular}{cccllllllll}
\toprule
\multicolumn{2}{c}{Method} & \phantom{.}   &  \multicolumn{2}{c}{Function $f_1$}  &  \phantom{.} & \multicolumn{2}{c}{Function $f_2$}\phantom{.} && \multicolumn{2}{c}{Function $f_3$}\\
\cmidrule{1-2} \cmidrule{4-5}  \cmidrule{7-8} \cmidrule{10-11}
Method&Activation&& $(d,M)=(5,300)$& $(d,M)=(10,500)$ && $(d,M)=(5,500)$ & $(d,M)=(10,1000)$ && $(d,M)=(5,500)$ & $(d,M)=(10,200)$\\
\midrule
\multirow{2}{*}{Fully Sampled}&\multirow{2}{*}{Fourier}&&$\num{1.41e-3}$&$\num{1.55e-2}$&&$\num{4.01e-2}$&$\num{3.73e-2}$&&$\num{4.89e-0}$&$\num{5.70e-0}$\\
&&&($\lambda=\num{1e-4}$)&($\lambda=\num{1e-3}$)&&($\lambda=\num{1e-3}$)&($\lambda=\num{1e-4}$)&&($\lambda=0$)&($\lambda=\num{1e-1}$)\\\midrule
\multirow{2}{*}{Noisy gradients}&\multirow{2}{*}{Fourier}&&$\num{1.03e-2}$&$\num{7.56e-3}$&&$\num{7.04e-3}$&$\num{4.16e-3}$&&$\num{1.38e-2}$&$\num{7.83e-0}$\\
&&&($\lambda=0$)&($\lambda=\num{1e-2}$)&&($\lambda=0$)&($\lambda=\num{1e-2}$)&&($\lambda=\num{1e-4}$)&($\lambda=\num{1e-4}$)\\
\bottomrule
\end{tabular}
}
    \label{tab:results_ablations}
\end{table}

\section{Implementation Details}\label{app:impl_details}

We optimize the loss functions for GFT and for the feature refinement with the Adam optimizer using a learning rate of $\num{1e-4}$ for $40000$ steps.
The regularization $\epsilon$ for solving the least squares problem \eqref{eq:b_von_w} is set to $\epsilon=\num{1e-7}$. 
For the neural network optimization, we use the Adam optimizer with a learning rate of $\num{1e-3}$ for $100000$ steps.
In all cases, we discretize the spatial integral for the regularization term in \eqref{eq:EmpRiskReg} by $1000$ samples.
For the kernel ridge regression, we use a Gauss kernel with its parameter chosen by the median rule.
That is, we set it to the median distance of two points in the dataset.
The PyTorch implementation corresponding to our experiments is available at \url{https://github.com/johertrich/generative_feature_training}.

\end{document}